\documentclass{article}

\usepackage[preprint]{neurips_2026}
\usepackage[utf8]{inputenc} 
\usepackage[T1]{fontenc}    
\usepackage{hyperref}       
\usepackage{url}            
\usepackage{booktabs}       
\usepackage{amsfonts}       
\usepackage{nicefrac}       
\usepackage{microtype}      
\usepackage{xcolor}         
\usepackage{siunitx}
\usepackage{graphicx}
\usepackage{subcaption} 
\usepackage{amsmath}
\usepackage{amssymb}
\usepackage{multirow}
\usepackage{algorithm}
\usepackage{algpseudocode}
\usepackage{wrapfig}

\title{Hyperspherical Autoencoder for High-Fidelity Image Reconstruction and Generation}

\author{%
  Hun Chang\thanks{Equal contribution.} \\
  KAIST AI \\
  \texttt{hun.mark.chang@kaist.ac.kr} \\
  \And
  Byunghee Cha\footnotemark[1] \\
  KAIST AI \\
  \texttt{paulcha1025@kaist.ac.kr} \\
  \And
  Jong Chul Ye \\
  KAIST AI \\
  \texttt{jong.ye@kaist.ac.kr} \\
}

\begin{document}

\maketitle

\begin{figure}[h]
    \centering
    \includegraphics[width=1.0\linewidth]{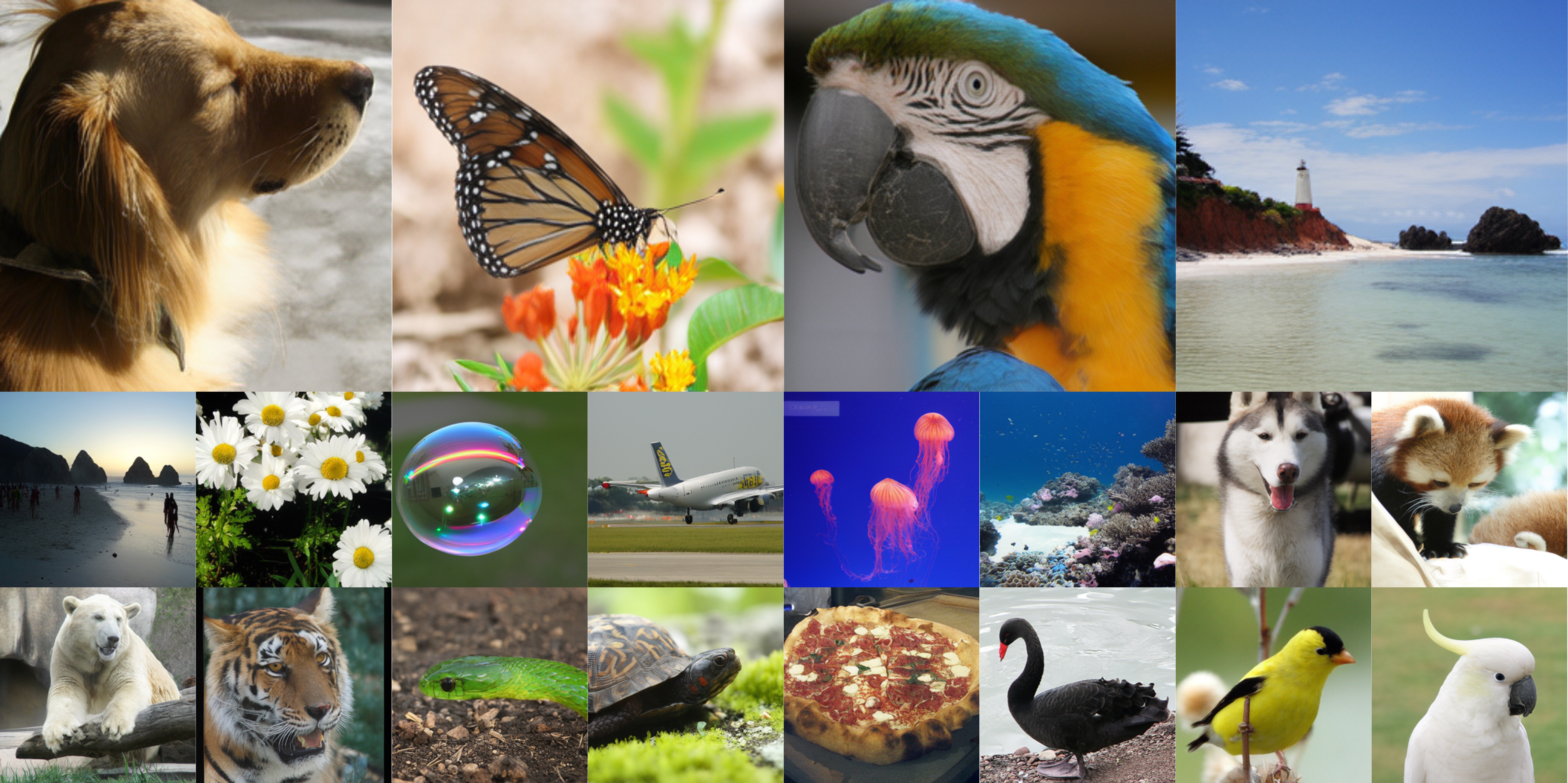}
    \caption{Representative high-fidelity reconstruction and generation samples from HAE.}
    \label{fig:intro_samples}
\end{figure}

\begin{abstract}
  Recent studies have explored using pretrained Vision Foundation Models (VFMs) such as DINO for generative autoencoders, showing strong generative performance. Unfortunately, existing approaches often suffer from limited reconstruction fidelity due to the loss of high-frequency details. In this work, we present the \textbf{\em Hyperspherical Autoencoder (HAE)}, a framework that bridges semantic representation and pixel-level reconstruction. Our key insight is that while semantic information in contrastive representations is primarily directional, enforcing strict magnitude matching hinders the preservation of fine-grained details. To address this, we introduce a {\em Directional Feature Alignment} objective that enforces semantic consistency while allowing flexible feature magnitudes for detail retention, alongside a {\em Hierarchical Convolutional Patch Embedding} module to enhance local structure preservation. Furthermore, observing that SSL-based representations intrinsically lie on a hypersphere, we employ {\em Riemannian Flow Matching} to train a Diffusion Transformer (DiT) directly on this spherical latent manifold. Notably, our manifold-aware DiT exhibits highly efficient convergence, achieving an exceptional gFID of \textbf{1.96} alongside a reconstruction rFID of \textbf{0.78} and a PSNR of \textbf{25.2} dB, validating the advantages of our manifold-aware approach. Code is available at \url{https://github.com/wkdgnsgo/HAE}.
\end{abstract}

\section{Introduction}
\label{sec:intro}

Diffusion models~\citep{ho2020ddpm,song2021ddim,Peebles2022DiT} have revolutionized image generation, and a key factor behind their success is the use of latent diffusion model (LDM)~\citep{rombach2022ldm}, where generation is performed in the compact latent space of a pretrained autoencoder. Consequently, the design of the latent space plays a decisive role in determining the overall performance of diffusion models. Recent work has shown that aligning internal diffusion representations with pretrained Vision Foundation Models (VFMs) can accelerate convergence and improve semantic generation, suggesting that similar benefits may be obtained by incorporating VFM representations into latent autoencoders. Along this direction, methods such as RAE~\citep{zheng2025rae} leverage pretrained VFMs like DINOv2~\citep{oquab2023dinov2} as encoders, enabling the latent space to capture richer semantic information. However, this comes with a critical trade-off: although VFM-based autoencoders improve semantic expressiveness, they often exhibit substantially degraded pixel-level reconstruction fidelity, e.g., lower PSNR, compared to standard VAEs.

We attribute this limitation to two primary factors: the aggressive downsampling in standard Vision Transformer (ViT) patch embeddings and the rigidity of feature alignment objectives. Existing methods typically employ Mean Squared Error (MSE) to align the student encoder with the VFM teacher. However, strictly enforcing both magnitude and directional alignment creates a \textit{gradient conflict} between semantic preservation and pixel reconstruction. This over-constrained optimization landscape prevents the encoder from learning the subtle high-frequency features necessary for fidelity.

To address these challenges, we introduce the \textit{Hyperspherical Autoencoder (HAE)}, a framework designed to reconcile semantic abstraction with high-fidelity reconstruction. Through systematic analysis, we discover that the semantic integrity of VFM representations is predominantly encoded in their directional component, while magnitude plays a negligible role. Driven by this empirical finding, we adopt a \textit{Hierarchical Convolutional Patch Embedding} to preserve local texture information, paired with a \textit{Directional Feature Alignment} strategy using cosine similarity. Unlike MSE, our objective relaxes the magnitude constraint, providing the encoder with the necessary degrees of freedom to minimize reconstruction error while prioritizing semantic alignment via feature direction. 

Building upon the insight that directional information dictates semantics, we extend this geometric perspective to the generative stage. Theoretically and empirically, representations derived from Self-Supervised Learning (SSL, e.g., DINO) naturally reside on a hyperspherical manifold rather than a flat Euclidean space. Consequently, applying standard Euclidean diffusion to these latents introduces a fundamental geometric mismatch. To resolve this, we model the generative dynamics directly on the hypersphere via \textbf{Riemannian Flow Matching (RFM)}~\citep{chen2023riemannianfm}. By constraining the generative process to the spherical manifold and modeling geodesics, we eliminate redundant radial variations and focus solely on semantically meaningful angular dynamics, providing a geometrically principled and highly efficient formulation.
Our contributions can be summarized as follows:

\begin{itemize}
    \item We propose a \textbf{Hierarchical Convolutional Patch Embedding} that mitigates the information bottleneck of standard ViT patchification, significantly enhancing local detail preservation.
    \item We introduce \textbf{Directional Feature Alignment}, which alleviates the optimization conflict inherent in MSE-based distillation. This enables the encoder to simultaneously achieve high semantic alignment and highly competitive reconstruction fidelity (25.2 dB PSNR).
    \item We empirically reveal that the semantic properties of VFM latents are strictly governed by their directional components. Guided by this geometric insight and the intrinsic spherical nature of SSL features, we formulate the generative process via \textbf{Riemannian Flow Matching}. This manifold-aware approach resolves the geometric mismatch of Euclidean baselines, significantly accelerating training convergence and enhancing generation quality.
\end{itemize}

\section{Related Work}

\paragraph{Representation Alignment.}
Recent work has shown that leveraging semantic representations can significantly improve the training efficiency of diffusion-based generative models. Following the introduction of REPA~\citep{yu2025repa}, which aligns intermediate features of DiT with pretrained DINOv2 representations, several extensions have been proposed. REG~\citep{wu2025representation} further enhances this paradigm by introducing a class token into DiT and aligning it with the DINOv2 class token in feature space during generation, leading to even faster convergence. Beyond external representation models, a number of self-contained approaches have demonstrated similar efficiency gains: SRA~\citep{jiang2025sra} aligns intermediate features at higher noise levels with later-layer features at lower noise levels, while LSEP~\citep{yun2025lsep} shows that merely enforcing linear separability of intermediate features via a classification probe can improve convergence. REPA-E~\citep{leng2025repae} extends the range of representation alignment training to the Latent Autoencoder itself, jointly optimizing VAE parameters under the REPA objective, further improving convergence speed.

\paragraph{Semantic Latent Autoencoders.}
In parallel, substantial effort has been devoted to improving latent autoencoders by aligning their latent spaces with semantic representations. Methods such as VA-VAE~\citep{yao2025vavae} and MAETok~\citep{chen2025maetok} explicitly align VAE latents with pretrained vision foundation models, achieving competitive or even improved rFID while dramatically accelerating gFID convergence. RAE~\citep{zheng2025rae} takes a more direct approach by replacing the VAE encoder entirely with a pretrained vision foundation model, eliminating the need for explicit alignment losses and fully exploiting pretrained representations. However, RAE and similar VFM-encoder-based tokenizers exhibit significantly degraded pixel-level reconstruction quality, especially suffering low PSNR. This limitation arises because pretrained vision foundation models prioritize semantic abstraction over fine-grained details, which are essential for accurate reconstruction. We identify the key architectural bottleneck causing this information loss and introduce a minimal modification that improves reconstruction fidelity while preserving the semantic strengths of VFM-based encoders.


\paragraph{Riemannian Flow Matching.}
Chen and Lipman~\cite{chen2023riemannianfm} extended the idea of flow matching (FM)~\citep{lipman2022flow} beyond Euclidean spaces and proposed \emph{Riemannian Flow Matching} (RFM), enabling Continuous Normalizing Flow (CNF) training on general Riemannian manifolds.
$\mathcal{M}$. 
In this setting, the model learns a time-dependent tangent vector field
$\mathbf{v}_\theta(\mathbf{x},t) \in T_{\mathbf{x}}\mathcal{M}$ 
that transports samples from a base distribution $p$ to the target data distribution $q$.

Specifically, the RFM objective is defined as the regression loss
\begin{equation}
\mathcal{L}_{\mathrm{RFM}}(\theta)
=
\mathbb{E}_{t,\,\mathbf{x}_t} 
\Big[
\|
\mathbf{v}_{\theta}(\mathbf{x}_t,t)
-
\mathbf{u}(\mathbf{x}_t,t)
\|_g^{2}
\Big],
\end{equation}
where $\|\cdot\|_g$ denotes the norm induced by the Riemannian metric,
and $\mathbf{u}(\mathbf{x}_t,t)$ is the marginal target flow generating a probability path 
between $p$ and $q$.

Following the conditional flow matching, the target vector field can be expressed as the marginalization
of a conditional flow $\mathbf{u}_t(\mathbf{x}\mid \mathbf{x}_1)$:
\begin{equation}
\mathbf{u}(\mathbf{x}_t,t)
=
\int_{\mathcal{M}}
\mathbf{u}_t(\mathbf{x}_t \mid \mathbf{x}_1)\,
\frac{p_t(\mathbf{x}_t \mid \mathbf{x}_1)\,q(\mathbf{x}_1)}{p_t(\mathbf{x}_t)}
\,d\mathrm{vol}_{\mathbf{x}_1}.
\end{equation}
A key distinction from standard FM lies in the design of the conditional flow on manifolds.
RFM introduces the notion of a \emph{premetric} $d(\mathbf{x},\mathbf{x}_1)$ and constructs
a conditional trajectory $\psi_t(\mathbf{x}_0\mid \mathbf{x}_1)$ that monotonically decreases
this distance:
\begin{equation}
d(\psi_t(\mathbf{x}_0\mid \mathbf{x}_1),\mathbf{x}_1)=\kappa(t)\,d(\mathbf{x}_0,\mathbf{x}_1),
\end{equation}
where $\kappa(t)$ is a decreasing scheduler satisfying $\kappa(0)=1$ and $\kappa(1)=0$.
Under this formulation, the conditional vector field admits the closed-form expression
\begin{equation}
\mathbf{u}_t(\mathbf{x}\mid \mathbf{x}_1)
=
\frac{d \log \kappa(t)}{dt}\,
d(\mathbf{x},\mathbf{x}_1)\,
\frac{\nabla d(\mathbf{x},\mathbf{x}_1)}
{\|\nabla d(\mathbf{x},\mathbf{x}_1)\|_g^{2}}.
\end{equation}
In the special case where the premetric is chosen as the geodesic distance $d_g$,
the resulting conditional flow reduces to the constant-speed geodesic connecting the endpoints,
which can be expressed analytically using the exponential and logarithmic maps:
\begin{equation}
\psi_t(\mathbf{x}_0,\mathbf{x}_1)
=
\exp_{\mathbf{x}_1}(\kappa(t)\log_{\mathbf{x}_1}(\mathbf{x}_0)).
\end{equation}
This construction makes RFM simulation-free on simple manifolds, while remaining scalable to general
geometries.

\begin{figure}[t!]
    \centering
    \begin{minipage}{0.48\textwidth}
        \centering
        \includegraphics[width=\linewidth]{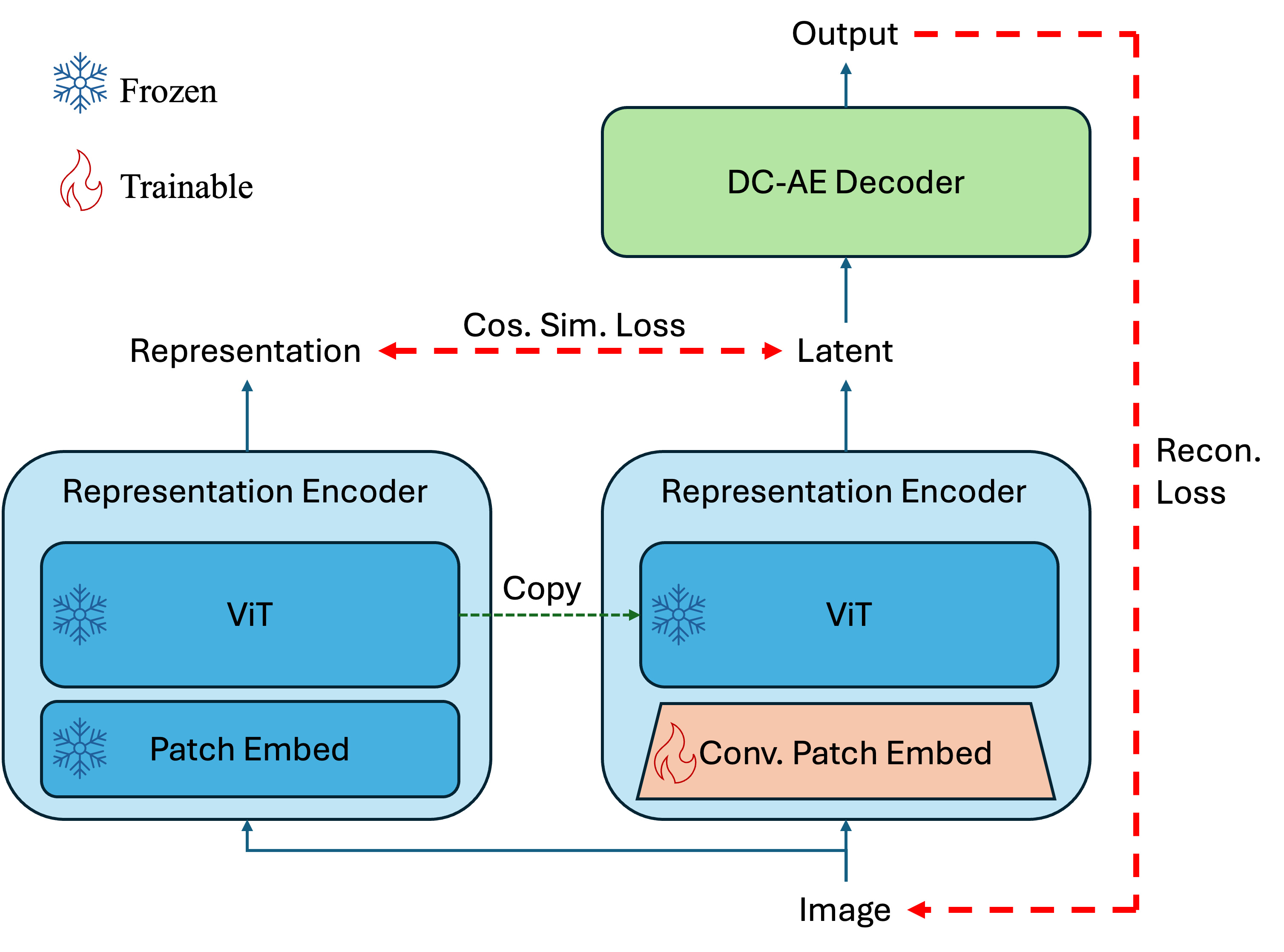}
        \caption{HAE architecture and training loss overview}
        \label{fig:architecture}
    \end{minipage}
    \hfill
    \begin{minipage}{0.48\textwidth}
        \centering
        \captionof{table}{Comparison of autoencoder models on ImageNet 256$\times$256.}
        \resizebox{0.6\linewidth}{!}{
        \begin{tabular}{lcc}
            \toprule
            \textbf{Model} & \textbf{rFID} & \textbf{PSNR} \\
            \midrule
            SD-VAE & 0.62 & 26.04 \\
            RAE & 0.59 & 18.94 \\
            MAETok & 0.48 & 23.61 \\
            VAVAE & 0.28 & 27.96 \\
            \midrule
            HAE (Ours) & 0.78 & 25.20 \\
            \bottomrule
        \end{tabular}}
        \label{tab:main_results}
        
        \vspace{1.5em} 
        
        \captionof{table}{Effect of latent smoothing.}
        \resizebox{\linewidth}{!}{
        \begin{tabular}{lccc}
            \toprule
            Setting & rFID & gFID  & PSNR \\
            \midrule
            \textit{w/o} Latent Smoothing & 0.37 & 32.64 & 26.2 \\
            \textit{w/} Latent Smoothing & 0.78 & 2.65 & 25.2 \\
            \bottomrule
        \end{tabular}}
        \label{tab:ablation}
    \end{minipage}
    \vspace{-10pt}
\end{figure}

\section{Hyperspherical Autoencoder (HAE)}
\label{sec:method}

Our {Hyperspherical Autoencoder} (HAE) is  designed to equip pre-trained vision foundation models with high-fidelity reconstruction capabilities. Our approach addresses the information bottleneck in standard Vision Transformers (ViTs) and introduces a scale-decoupled alignment strategy to preserve high-frequency details without compromising semantic integrity.
The overall architecture  and its training objective is illustrated in Fig.~\ref{fig:architecture},
and the details are provided below.

\subsection{Architecture}

\noindent\textbf{Enhanced Patch Embedding.}
Standard ViT architectures, including DINOv3~\cite{simeoni2025dinov3}, typically implement patch embedding using a single convolutional layer. We hypothesize that this aggressive, non-overlapping downsampling operation acts as a primary bottleneck, causing the irreversible loss of high-frequency spatial details essential for reconstruction tasks.

To mitigate this, we replace the standard single-layer embedding with a \textit{Hierarchical Convolutional Stem}. Specifically, we design a four-stage Convolutional Neural Network (CNN) that progressively downsamples the input image. This enhanced patch embedding layer allows the model to capture fine-grained local features (e.g., edges, textures) in the early stages, which are typically discarded by the large-kernel convolution of standard ViTs. The output of this module matches the dimension of the Transformer blocks, serving as a rich input token sequence $\mathbf{z}_{0}$. For the subsequent encoding stages, we utilize the pre-trained DINOv3 Transformer backbone, keeping all parameters frozen to preserve the original semantic representations.

\noindent\textbf{Decoder Architecture.}
For the decoding stage, we adopt the lightweight yet effective decoder architecture proposed in DC-AE~\cite{chen2024deep}. This decoder is designed to efficiently upsample the latent tokens $\mathbf{z}$ back to the pixel space $\hat{\mathbf{x}}$ while minimizing computational overhead. Detailed architectural configurations are provided in Appendix~\ref{sec:appendix_autoencoder}.

\subsection{Training}

\noindent\textbf{Directional feature alignment.} Standard feature alignment via Mean Squared Error (MSE) strictly enforces both magnitude and directional matching. However, matching the exact magnitude of texture-invariant VFM features creates an over-constrained gradient conflict that degrades pixel-level reconstruction~\cite{chen2025aligning, tang2025unilip}. To alleviate this, we adopt \textit{directional feature alignment} using cosine similarity: $\mathcal{L}_{\text{align}} = 1 - (\mathbf{z}_{S} \cdot \mathbf{z}_{T}) / (\|\mathbf{z}_{S}\|_2 \|\mathbf{z}_{T}\|_2)$, where $\mathbf{z}_{S}$ and $\mathbf{z}_{T}$ denote the student and teacher features, respectively. This isolates semantic alignment to the angular component, leaving feature magnitudes flexible for high-fidelity reconstruction. This relaxation resolves the reconstruction-alignment trade-off and motivates our spherical generation formulation in Section~\ref{sec:rfm}.

\begin{figure}[t!]
    \captionsetup{font=footnotesize}
    \centering
    \begin{minipage}{0.48\textwidth}
        \centering
        \includegraphics[width=\linewidth]{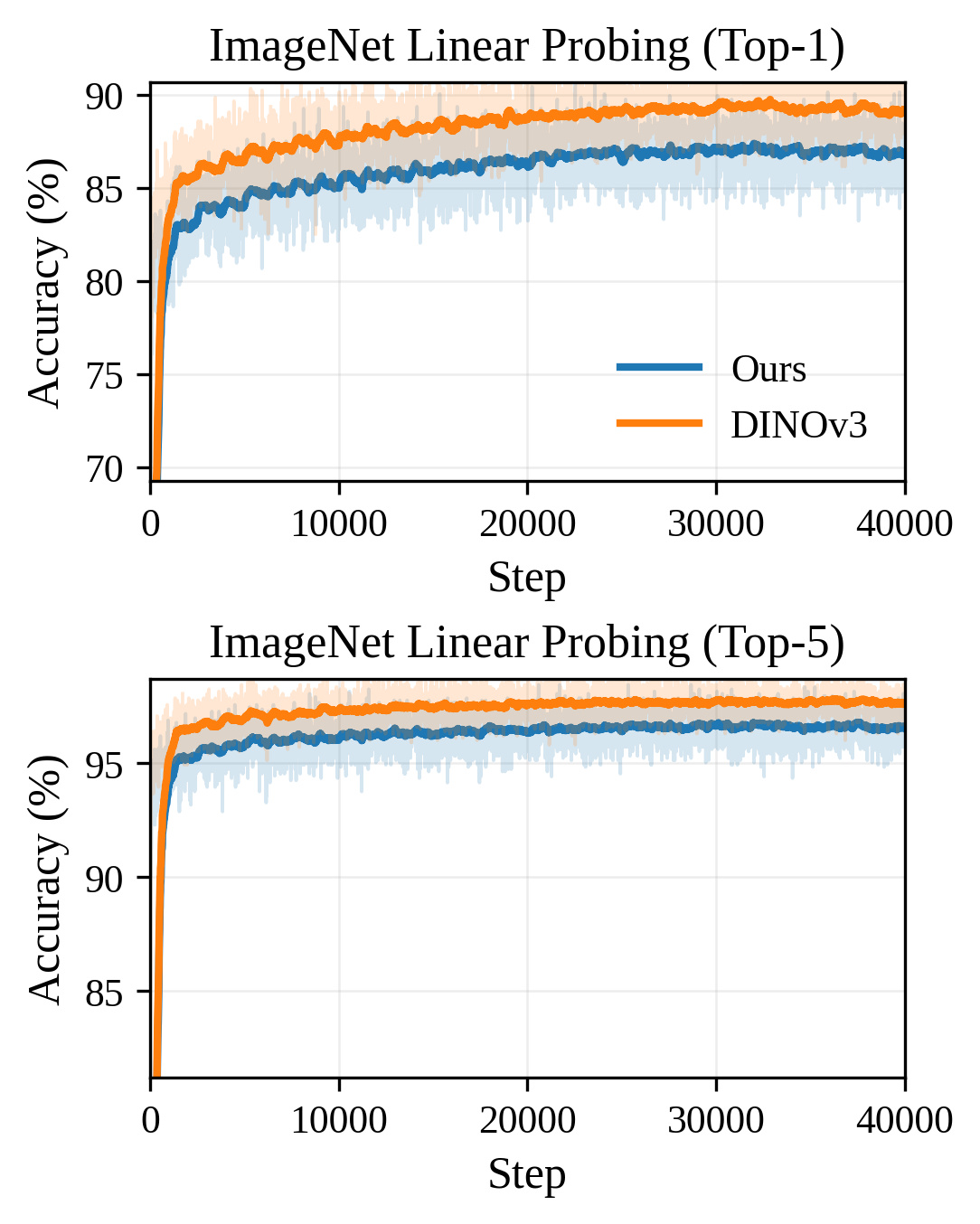}
        \caption{Linear probing results on ImageNet-1K. HAE retains robust semantic information throughout training.}
        \label{fig:linear_probing}
    \end{minipage}
        \hfill
    \begin{minipage}{0.48\textwidth}
        \centering
        \includegraphics[width=\linewidth]{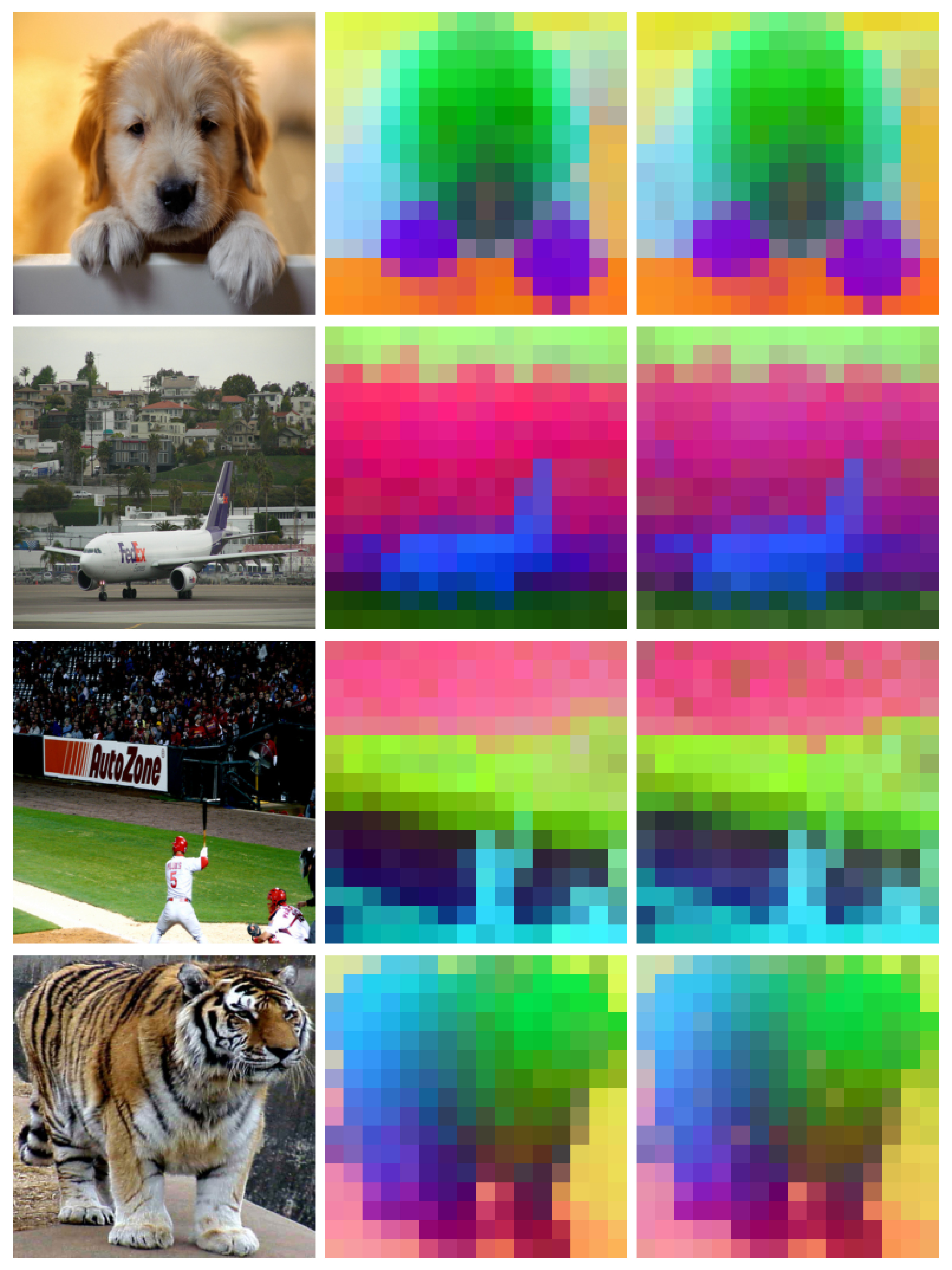}
        \caption{PCA visualization of feature representations. HAE faithfully preserves the teacher's semantic structure.}
        \label{fig:pca_analysis}
    \end{minipage}
    \vspace{-10pt}
\end{figure}

We empirically verify that this relaxation preserves semantic integrity. Linear probing on ImageNet-1K (Figure~\ref{fig:linear_probing}) shows HAE retains competitive accuracy (87\% Top-1, 97\% Top-5) compared to DINOv3 (89\%, 98\%). Furthermore, PCA visualizations of the feature maps (Figure~\ref{fig:pca_analysis}) confirm that the student faithfully reproduces the teacher’s fine-grained semantic geometry. These results demonstrate that directional alignment effectively transfers semantic priors while supporting high-fidelity reconstruction.

\noindent\textbf{Progressive Training Strategy.}
We train our model in four progressive stages to balance semantic alignment and reconstruction fidelity.

\textit{Stage 1: Semantic-Structural Alignment.}
In the initial phase, we train the encoder—specifically focusing on the hierarchical patch embedding—alongside the decoder to establish a stable latent space. We employ a combination of the directional alignment loss, pixel-wise reconstruction loss, and perceptual loss:
\begin{equation}
    \mathcal{L}_{\text{Stage1}} = \lambda_{\text{cos}}\mathcal{L}_{\text{align}} + \lambda_{\text{L1}}\|\mathbf{x} - \hat{\mathbf{x}}\|_1 + \lambda_{\text{lpips}}\mathcal{L}_{\text{LPIPS}}(\mathbf{x}, \hat{\mathbf{x}})
\end{equation}
where $\mathcal{L}_{\text{LPIPS}}$ denotes the LPIPS loss for perceptual quality. Notably, this configuration—combining the hierarchical patch embedding with the cosine alignment objective—maximizes the model's pure reconstruction capacity, yielding exceptionally high PSNR and low rFID scores prior to any generative regularization.

\textit{Stage 2: Adversarial Adaptation with Stochastic Latent Smoothing.}
From this stage onwards, we freeze the entire encoder to preserve the established semantic latent space. To bridge the domain gap between strict reconstruction and downstream generative sampling, we introduce adversarial training coupled with a stochastic latent noise injection strategy.

Following RAE~\cite{zheng2025diffusion}, we inject isotropic Gaussian noise into the latent representations with a randomized continuous scale:
$\mathbf{z}' = \mathbf{z} + \sigma \boldsymbol{\epsilon}$, where $\sigma \sim \mathcal{U}(0, \tau)$ is sampled per batch and $\boldsymbol{\epsilon} \sim \mathcal{N}(\mathbf{0}, \mathbf{I})$.
Rather than constraining latents to the exact hyperspherical surface, this perturbation thickens the latent manifold by forming a local volumetric neighborhood around each semantic direction.
Training the decoder to reconstruct from these perturbed latents encourages local smoothness and improves robustness to imperfect latent trajectories produced by the Diffusion Transformer (DiT) during inference.




This smoothing introduces a trade-off between pure reconstruction fidelity and downstream generative quality. As shown in Table~\ref{tab:ablation}, \textbf{training without latent smoothing} achieves stronger reconstruction metrics, with an rFID of \textbf{0.37} and a PSNR of \textbf{26.2} dB, but severely degrades generation quality, yielding a gFID above \textbf{32.64}. In contrast, \textbf{training with latent smoothing} slightly sacrifices reconstruction fidelity, yielding an rFID of \textbf{0.78} and a PSNR of \textbf{25.2} dB, but substantially improves generative performance to an impressive gFID of \textbf{2.65}. This demonstrates that latent smoothing is an essential and highly effective regularizer for enabling generative sampling while preserving competitive reconstruction quality.

Coupled with this stochastic smoothing, we incorporate an adversarial loss~\cite{sauer2023stylegan, zheng2025diffusion} to further enhance the overall reconstruction quality and compensate for potential smoothing artifacts. The final objective is formulated as:
\begin{equation}
    \mathcal{L}_{\text{Stage2}} = \mathcal{L}_{\text{Stage1}} + \lambda_{\text{adv}}\mathcal{L}_{\text{GAN}}(\mathbf{x}, \hat{\mathbf{x}})
\end{equation}
Detailed training configurations, including the specific discriminator architecture, are provided in Appendix~\ref{sec:appendix_autoencoder}.

\begin{figure}[t!]
    \centering
    \captionsetup{font=small}
    \includegraphics[width=1.0\linewidth]{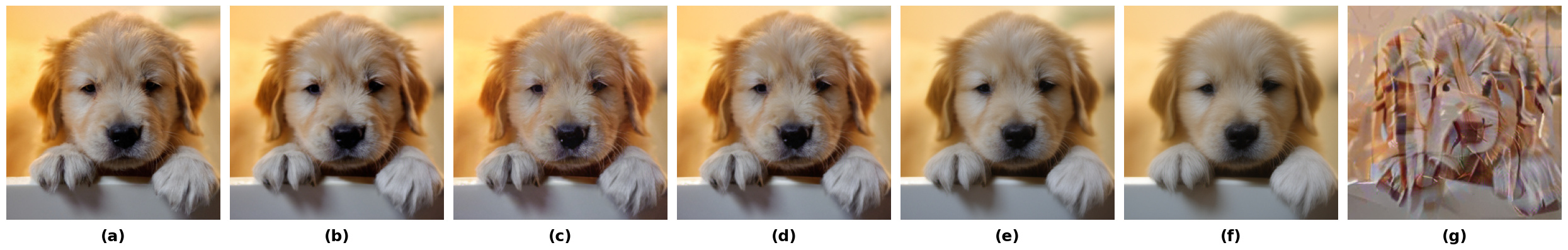}
    \caption{\textbf{Magnitude robustness and directional sensitivity of HAE.}
    (a) Ground Truth (GT). (b) Original reconstruction.
    (c--f) Reconstructions after fixing latent magnitudes to $s=8, 10, 16,$ and $20$.
    (g) Reconstruction after injecting angular noise ($g=0.1$) while preserving the original magnitude.
    The contrast between stable magnitude-rescaled reconstructions and degraded angle-perturbed reconstruction suggests that semantic and structural information is predominantly encoded in feature direction.}
    \label{fig:mag_comparisons}
    \vspace{-10pt}
\end{figure}

\subsection{Image Generation via HAE}
\label{sec:rfm}

Through empirical analysis, we observe that the reconstruction output of our trained autoencoder is primarily governed by the \emph{direction} of latent features, being relatively insensitive to their \emph{magnitude} as long as the values remain within a reasonable numerical bound.


To systematically validate this property, we rescale the latent representation $\mathbf{z}$ extracted by the encoder to fixed magnitudes along the channel dimension. Specifically, for each spatial location $(i,j)$, we compute
$\tilde{\mathbf{z}}_{ij}^{(s)} = s \cdot \frac{\mathbf{z}_{ij}}{\|\mathbf{z}_{ij}\|_2},\quad s \in \{8, 10, 16, 20\}.$

As illustrated in Figure~\ref{fig:mag_comparisons}(c-f), the reconstructed images maintain high visual fidelity and semantic consistency despite large variations in the imposed scale, confirming the decoder's strong robustness to magnitude shifts.

Conversely, to evaluate directional sensitivity, we inject a small Gaussian noise $\epsilon \sim \mathcal{N}(0, g^2\mathbf{I})$ (with $g=0.1$) strictly into the normalized directional vector. We re-normalize this perturbed vector and multiply it by the original magnitude $\|\mathbf{z}\|_2$ to ensure the latent scale remains unchanged, decoding $\hat{\mathbf{z}}_{\text{noise}} = \|\mathbf{z}\|_2 \cdot \frac{\bar{\mathbf{z}} + \epsilon}{\|\bar{\mathbf{z}} + \epsilon\|_2}$. As shown in Figure~\ref{fig:mag_comparisons}(g), injecting even a minimal amount of angular noise catastrophically degrades the image quality. 

This stark contrast provides empirical proof that the core semantic and structural information in our latent space is predominantly encoded within the directional (angular) component. Motivated by this foundational property, we propose to perform Riemannian Flow Matching~\citep{chen2023riemannianfm} strictly on a patch-wise spherical manifold.

Formally, we model the latent space as a product manifold of $N$ hyperspheres:
\begin{equation}
\mathcal{M} = \underbrace{S_R^{C} \times S_R^{C} \times \cdots \times S_R^{C}}_{N\ \text{times}}, \quad \text{where} \quad S_R^{C} = \left\{ \mathbf{z} \in \mathbb{R}^{C} \;\middle|\; \lVert \mathbf{z} \rVert_2 = R \right\}.
\end{equation}
Each patch-level latent $\mathbf{x}^{(i)} \in \mathbb{R}^C$ is therefore constrained to lie on a sphere of fixed radius $R$. Instead of linear interpolation in Euclidean space, we define the interpolation path between $\mathbf{x}_0$ and $\mathbf{x}_1$ along the geodesic of the spherical manifold. For each patch index $i$, the geodesic interpolation at time $t \in [0,1]$ is given by
\begin{equation}
\mathbf{x}_t^{(i)} = \frac{\sin\!\big((1 - t)\Omega^{(i)}\big)}{\sin\!\big(\Omega^{(i)}\big)} \, \mathbf{x}_0^{(i)} + \frac{\sin\!\big(t \Omega^{(i)}\big)}{\sin\!\big(\Omega^{(i)}\big)} \, \mathbf{x}_1^{(i)},
\end{equation}
where the angular distance $\Omega^{(i)}$ is defined as
\begin{equation}
\Omega^{(i)} = \arccos\!\left( \frac{\langle \mathbf{x}_0^{(i)}, \mathbf{x}_1^{(i)} \rangle}{R^2} \right).
\end{equation}
The target velocity $\mathbf{u}_t^{(i)}$ is obtained by differentiating the geodesic with respect to time:
\begin{equation}
\mathbf{u}_t^{(i)} = \frac{d}{dt}\,\mathbf{x}_t^{(i)} = \frac{\Omega^{(i)}}{\sin\!\big(\Omega^{(i)}\big)} \left( \cos\!\big(t\Omega^{(i)}\big)\,\mathbf{x}_1^{(i)} - \cos\!\big((1 - t)\Omega^{(i)}\big)\,\mathbf{x}_0^{(i)} \right).
\end{equation}

In practice, while the target velocity $\mathbf{u}_t^{(i)}$ is mathematically bound to the tangent space of the hypersphere, the neural network $\mathbf{v}_\theta(\mathbf{x}_t, t)$ outputs unconstrained vectors in the Euclidean space $\mathbb{R}^C$. Although the trained decoder is highly robust to magnitude variations, allowing the neural network to freely predict radial velocities introduces unnecessary variance and wastes the model's capacity on semantically negligible magnitude shifts.

To explicitly enforce the manifold constraint and ensure training stability, we decompose the network's prediction into tangent and radial components. We then apply a Riemannian Flow Matching objective on the tangent space, alongside a \textit{radial penalty} to strictly suppress any velocity orthogonal to the manifold. As demonstrated in Table~\ref{tab:lambda_rad_ablation}, calibrating this penalty is crucial for optimal performance; setting $\lambda_{\text{rad}}=0.1$ effectively suppresses radial drift while maintaining generative quality, significantly outperforming unconstrained ($\lambda_{\text{rad}}=0.0$) or overly restricted ($\lambda_{\text{rad}}=1.0$) settings. The overall objective is formulated as:
\begin{equation}
\mathcal{L}(\theta) = \mathbb{E}_{t,\,\mathbf{x}_0,\,\mathbf{x}_1} \left[ \frac{1}{N} \sum_{i=1}^{N} \left( \left\| \mathbf{v}_{\text{tan}}^{(i)} - \mathbf{u}_t^{(i)} \right\|_2^{\,2} + \lambda_{\text{rad}} \left\| \mathbf{v}_{\text{rad}}^{(i)} \right\|_2^{\,2} \right) \right],
\end{equation}
where the radial component is computed as $\mathbf{v}_{\text{rad}}^{(i)} = \langle \mathbf{v}_\theta^{(i)}, \bar{\mathbf{x}}_t^{(i)} \rangle \bar{\mathbf{x}}_t^{(i)}$ with the unit normal vector $\bar{\mathbf{x}}_t^{(i)} = \mathbf{x}_t^{(i)} / R$, and the projected tangent velocity is $\mathbf{v}_{\text{tan}}^{(i)} = \mathbf{v}_\theta^{(i)} - \mathbf{v}_{\text{rad}}^{(i)}$. The hyperparameter $\lambda_{\text{rad}}$ controls the strength of the penalty for deviating from the spherical manifold.

This formulation aligns the generative dynamics with the intrinsic geometry of the autoencoder latent space, focusing the model on directional variations and improving training efficiency and stability. Detailed pseudocode for training and inference is provided in Appendix~\ref{sec:appendix_algorithms}.

\section{Experiments}
\label{sec:experiments}

\begin{table*}[t]
\centering
\small
\setlength{\tabcolsep}{4pt} 
\renewcommand{\arraystretch}{1.15}
\caption{\textbf{Comparison of generative models on ImageNet 256$\times$256.}
We report gFID, IS, Precision, and Recall with and without classifier-free guidance (CFG).
Lower gFID is better, while higher IS, Precision, and Recall indicate better performance.}
\label{tab:imagenet256}

\resizebox{\linewidth}{!}{
\begin{tabular}{
l  
S[table-format=4.0] 
c 
*{4}{S[table-format=3.2]} 
*{4}{S[table-format=3.2]} 
}
\toprule
\textbf{Method} &
\textbf{Epochs} &
\textbf{\#Params} &
\multicolumn{4}{c}{\textbf{Generation@256 w/o guidance}} &
\multicolumn{4}{c}{\textbf{Generation@256 w/ guidance}} \\
\cmidrule(lr){4-7}\cmidrule(lr){8-11}
& & &
\textbf{gFID$\downarrow$} & \textbf{IS$\uparrow$} & \textbf{Prec.$\uparrow$} & \textbf{Rec.$\uparrow$} &
\textbf{gFID$\downarrow$} & \textbf{IS$\uparrow$} & \textbf{Prec.$\uparrow$} & \textbf{Rec.$\uparrow$} \\
\midrule

\multicolumn{11}{l}{\textbf{\itshape Autoregressive}} \\
\midrule
VAR~\citep{Tian2024VAR} & 350 & 2B & 1.92 & 323.1 & 0.82 & 0.59 & 1.73 & \bfseries 350.2 & 0.82 & 0.60 \\
MAR~\citep{li2024mar}  & 800 & 943M & 2.35 & 227.8 & 0.79 & 0.62 & 1.55 & 303.7 & 0.81 & 0.62 \\
\midrule

\multicolumn{11}{l}{\textbf{\itshape Euclidean Latent Diffusion \& Flow Matching}} \\
\midrule
DiT-XL/2~\citep{Peebles2022DiT}  & 400 & 675M & 9.62 & 121.5 & 0.67 & 0.67 & 2.27 & 278.2 & 0.83 & 0.57 \\
SiT-XL/2~\citep{ma2024sit}  & 400 & 675M &  8.61 & 131.7 & 0.68 & 0.67 & 2.06 & 270.3 & 0.82 & 0.59 \\
REPA~\citep{yu2025repa}  & 80 & 675M & 7.90 & 122.6 & 0.70 & 0.65 & { - } & { - } & { - } & { - } \\
  & 800 & 675M &  5.78 & 158.3 & 0.70 & 0.68 & { - } & { - } & { - } & { - } \\
DDT~\citep{wang2025ddt}  & 400 & 675M & 6.62 & 135.2 & 0.69 & 0.67 & 1.52 & 263.7 & 0.78 & 0.63 \\
VAVAE~\citep{yao2025vavae}  & 80 & 675M & 4.29 & { - } & { - } & { - } & { - } & { - } & { - } & { - } \\
  & 800 & 675M & 2.17 & 205.6 & 0.77 & 0.65 & { - } & { - } & { - } & { - } \\
MAETok~\cite{chen2025maetok} & 800 & 675M &  2.56 & 224.5 & { - } & { - } & 1.72 & 307.3 & { - } & { - }\\
REPA-E~\cite{leng2025repae} & 80 & 675M & 3.46 & 159.8 & 0.77 & 0.63 & { - } & { - } & { - } & { - } \\
 & 800 & 675M &  1.70 & 217.3 & 0.77 & 0.66 & { - } & { - } & { - } & { - } \\
RAE~\citep{zheng2025rae} &  &  &   &  &  &  &  &  &  & \\
\quad+LightningDiT-XL/1 (DINOv2-B)
  & 80 & \multirow{2}{*}{676M} & 4.28 & { - } & { - } & { - } & { - } & { - } & { - } & { - } \\
\quad+LightningDiT-XL/1 (DINOv2-S)
  & 800 &  & 1.87 & 209.7  & 0.80  & 0.63  & 1.41 & 309.4 & 0.80 & 0.63 \\
\quad+DiT$^{DH}$-XL/1  (DINOv2-B)
  & 80  & \multirow{2}{*}{879M} & 2.16 & 214.8 & 0.82 & 0.59 & { - } & { - } & { - } & { - } \\
  & 800 &  & 1.51 & 242.9  & 0.79 & 0.63 & 1.13 & 262.6 & 0.78 & 0.67 \\
\midrule

\multicolumn{11}{l}{\textbf{\itshape Riemannian Latent Flow Matching}} \\
\midrule
RAE (DINOv2-B)+LightningDiT-XL/1+RJF~\cite{Kumar2026LearningOM} & 80 & 677M & 3.62 & 186.2 & 0.82 & 0.52 & 2.81 & 201.22 & 0.82 & 0.56 \\[4pt]
HAE (DINOv3-L)+LightningDiT-XL/1 (Ours)
  & 80  & \multirow{2}{*}{677M} & 2.65 & 202.1 & 0.82 & 0.55 & { - } & { - } & { - } & { - } \\
  & 550 &  & 1.96 &  249.2  & 0.72  & 0.69  & 1.90 & 252.7 & 0.71 & 0.70 \\
\bottomrule
\end{tabular}
} 
\vspace{-2mm}
\end{table*}

\subsection{Image Generation via Riemannian Flow Matching}

\textbf{Implementation.}
For downstream image generation, we train the LightningDiT-XL architecture utilizing 8 NVIDIA B200 GPUs. All LightningDiT-XL experiments were conducted on 8 NVIDIA B200 GPUs, requiring approximately 20 minutes per epoch. The model is optimized for 550 epochs using AdamW with a learning rate of $2 \times 10^{-4}$, betas of $(0.9, 0.95)$, and a global batch size of 1024. 

Crucially, because our generative process is modeled via Riemannian Flow Matching (RFM) directly along the data-driven spherical manifold, the training dynamics converge significantly faster than standard Euclidean flow matching. As a consequence, we empirically observed that using a conventional, slow Exponential Moving Average (EMA) decay rate (e.g., 0.9999) creates a bottleneck; the EMA weights fail to swiftly track the rapidly optimizing parameters, leading to an unwarranted performance penalty. To correctly align the moving average with our accelerated convergence rate, we adopt a faster EMA decay rate of 0.9995. We generate 50,000 images using the Riemannian Euler sampler with manifold projection for 50 steps.

\textbf{Results.}
As shown in Table~\ref{tab:imagenet256}, HAE provides a remarkably effective latent space for downstream generative modeling. When paired with LightningDiT-XL, our model achieves a gFID of 2.65 without guidance, significantly outperforming competitive VAE-based foundations such as VAVAE (4.29) and the baseline RAE (4.28) under the exact same 80-epoch training budget and architecture (676M).

Notably, while RAE utilizes a larger architecture with specialized DDT heads (DiT$^{DH}$-XL, 879M) to achieve its optimal performance, our approach remains highly competitive using a standard, lighter LightningDiT-XL (676M). This highlights the parameter efficiency of our hyperspherical formulation: by restricting diffusion dynamics strictly to the directional manifold, the model effectively bypasses~the need to learn radial variations, achieving strong performance with fewer parameters.

\begin{wrapfigure}{r}{0.4\linewidth}
    \vspace{-8pt}
    \centering
    \includegraphics[width=\linewidth]{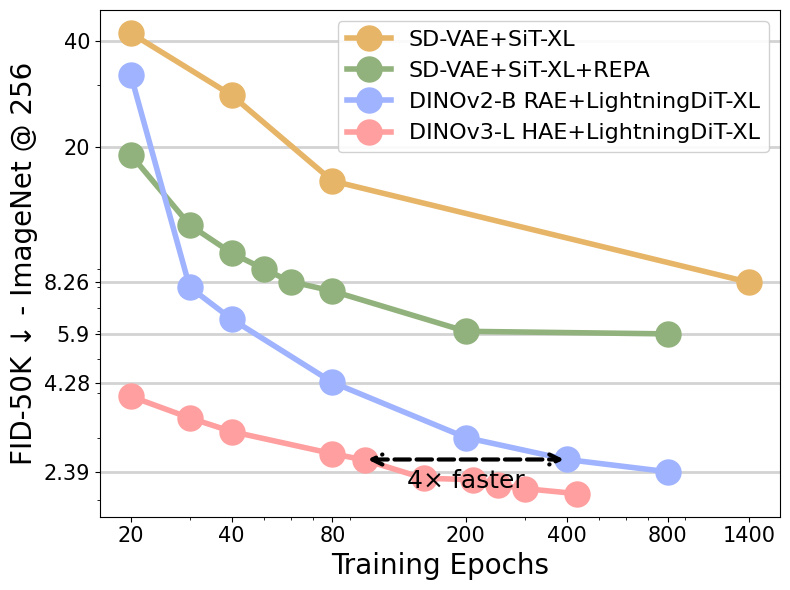}
    \caption{FID convergence comparison of various autoencoders and methods.}
    \label{fig:comparisons_speed}
    \vspace{-10pt}
\end{wrapfigure}

Figure ~\ref{fig:comparisons_speed} further highlights the training efficiency derived from our hyperspherical formulation. As explicitly shown in the gFID convergence curves, LightningDiT-XL trained on HAE latents exhibits an overwhelmingly faster convergence rate compared to other competitive baseline models. 
Crucially, while the baseline RAE employs standard Euclidean Flow Matching within an unconstrained latent space, our approach utilizes Riemannian Flow Matching to model diffusion dynamics strictly on the hyperspherical manifold. This exact geometric alignment ensures that the generative model does not waste representational capacity on arbitrary radial variations, allowing it to reach high-fidelity generation in a fraction of the typical training budget. Consequently, our model converges approximately 4$\times$ faster than the identical DiT architecture trained on RAE latents. Overall, these results demonstrate that respecting the intrinsic geometry of the latent space simultaneously improves generative quality and significantly accelerates training convergence.

\subsection{Ablation Studies}
\label{sec:ablation}

\begin{figure*}[t]
    \centering
    \captionsetup{font=small}
    
    \begin{minipage}[t]{0.32\textwidth}
        \vspace{0pt} 
        \centering
        \resizebox{0.9\linewidth}{!}{%
        \begin{tabular}{lccc}
            \toprule
            Method & gFID $\downarrow$ \\
            \midrule
            Euclidean FM  & 15.95 \\
            \textbf{Riemannian FM} & \textbf{2.65} \\
            \bottomrule
        \end{tabular}%
        }
        \captionof{table}{Ablation on generative dynamics: Euclidean Flow Matching (FM) vs. Riemannian Flow Matching (RFM).}
        \label{tab:rfm_vs_fm}
    \end{minipage}
    \hspace{0.03\textwidth}
    \begin{minipage}[t]{0.56\textwidth}
        \vspace{0pt} 
        \centering        
        \resizebox{0.9\linewidth}{!}{%
        \begin{tabular}{lcc}
            \toprule
            \textbf{Patch Embedding Strategy} & \textbf{rFID} $\downarrow$ & \textbf{PSNR} $\uparrow$ \\
            \midrule
            Standard ViT (Baseline) & 0.47 & 19.55 \\
            Hierarchical CNN (Ours) & \textbf{0.37} & \textbf{26.20} \\
            \bottomrule
        \end{tabular}}
        \captionof{table}{Ablation on patch embedding strategies. Utilizing the hierarchical CNN stem significantly improves high-frequency detail retention (PSNR). \textbf{Models in this table are evaluated without latent smoothing to isolate the pure architectural impact.}}
        \label{tab:ablation_patch}
    \end{minipage}
    \vspace{-12pt}
    
    
    
    
\end{figure*}

\paragraph{Effect of Hierarchical Convolutional Patch Embedding.}
To isolate the contribution of our modified patch embedding layer, we compare our HAE against a baseline that strictly uses the standard single-layer ViT patchification of the frozen DINOv3. As shown in Table~\ref{tab:ablation_patch}, avoiding modifications to the standard patch embedding results in a severe bottleneck for pixel-level reconstruction, yielding a PSNR of only 19.55 dB. In contrast, our Hierarchical CNN Patch Embedding effectively captures and preserves high-frequency spatial details early in the network. This architectural adjustment produces a massive improvement in reconstruction fidelity, boosting the PSNR to 19.55 dB while simultaneously improving the rFID.

\begin{wraptable}{r}{0.30\textwidth} 
\captionsetup{font=small}
    \vspace{-12pt}
    \centering
    \begin{tabular}{cc}
        \toprule
        $\lambda_{\text{rad}}$ & gFID $\downarrow$ \\
        \midrule
        0.0 & 4.40 \\
        1.0 & 5.70 \\
        0.1 & \textbf{3.90} \\
        \bottomrule
    \end{tabular}
    \caption{Ablation on radial penalty. Models in this study are trained for 20 epochs.}
    \label{tab:lambda_rad_ablation}
    \vspace{-12pt}
\end{wraptable}
\paragraph{Impact of the Radial Penalty.}
As summarized in Table~\ref{tab:lambda_rad_ablation}, we further investigate the role of the radial penalty term ($\lambda_{\text{rad}}$) during the early generative training phase. When $\lambda_{\text{rad}} = 0$, the network's velocity predictions are unconstrained, frequently drifting off the hyperspherical tangent space. This introduces unnecessary radial variance, leading to training instability. By enforcing a strict manifold constraint ($\lambda_{\text{rad}} > 0$), we explicitly suppress velocities orthogonal to the sphere, ensuring that the model focuses its capacity entirely on semantically meaningful directional shifts.

\paragraph{Riemannian vs. Euclidean Flow Matching.}
Finally, as reported in Table~\ref{tab:rfm_vs_fm}, we evaluate the necessity of Riemannian Flow Matching (RFM) compared to standard Euclidean Flow Matching (FM). To ensure a fair comparison, both models employ the LightningDiT-XL architecture and are evaluated at 80 epochs. Since the core structural and semantic information of our latent space is encoded in the feature direction (as proven in Section~\ref{sec:rfm}), standard Euclidean FM wastes modeling capacity on predicting arbitrary and redundant magnitude scales. By constraining the vector field to the geodesics of the spherical manifold, RFM inherently respects the geometry of the contrastive features. This results in more structurally coherent generation and a substantial improvement in performance, achieving an optimal gFID of 2.65 compared to the Euclidean baseline's gFID of 15.95.

\section{Conclusion}

We presented the Hyperspherical Autoencoder (HAE) to resolve the fundamental tension between semantic abstraction and pixel-level fidelity in Vision Foundation Models (VFMs). We identify the aggressive single-layer patch embedding of standard Vision Transformers as the primary bottleneck responsible for the irreversible loss of high-frequency details. 

To overcome this, HAE introduces a Hierarchical Convolutional Patch Embedding, successfully recovering and retaining crucial local textures. However, modifying this early stem inevitably shifts the feature space away from the pretrained teacher. To realign these representations without suffering from the over-constrained gradient conflict of standard mean-squared error (MSE), we propose Directional Feature Alignment via cosine similarity. This aligns the angular semantics with the pre-trained DINOv3 weights while granting the decoder the magnitude flexibility needed to reconstruct intricate structural details.

Recognizing that this magnitude-robust latent space inherently forms a hyperspherical manifold, we formulated the downstream generative dynamics via Riemannian Flow Matching (RFM). By projecting the vector field onto the tangent space and penalizing redundant radial variations, we prevent the model from wasting capacity on arbitrary magnitude scaling. Additionally, stochastic latent smoothing thickens this manifold, effectively bridging the domain gap between deterministic reconstruction and continuous generative sampling.

Evaluations on ImageNet-1K at $256 \times 256$ confirm our framework's superiority. HAE achieves a competitive reconstruction baseline of 0.78 rFID and 25.2 dB PSNR, empowering the downstream Diffusion Transformer to reach an optimal gFID of 1.96 with significantly accelerated convergence. Ultimately, HAE demonstrates that respecting the intrinsic geometry of the latent space—both architecturally and dynamically—provides a highly efficient and structurally coherent paradigm for next-generation visual synthesis.
\sisetup{
  detect-weight=true,
  detect-inline-weight=math,
  table-number-alignment=center
}

\medskip
\clearpage

{
\small
\bibliographystyle{unsrt}
\bibliography{ref} 
}

\clearpage
\appendix

\section{Technical Appendices and Supplementary Material}
\label{sec:appendix_algorithms}

In this appendix, we detail the training and inference procedures for Riemannian Flow Matching (RFM) on the HAE latent space. 

\subsection{Training Algorithm}
Algorithm~\ref{alg:training} outlines the training of our manifold-aware DiT. We strictly constrain the generative process to a hypersphere of radius $R=32.0$. Since our DINOv3-Large encoder inherently extracts $C=1024$ dimensional features, setting $R = \sqrt{2}\frac{\Gamma((C+1)/2)}{\Gamma(C/2)} \approx 32.0$ ensures unit variance per channel and optimal numerical scaling.

Furthermore, recent studies~\cite{teng2023relay, chen2023importance, hoogeboom2023simple, pmlr-v235-esser24a} indicate that high-dimensional latent spaces (e.g., RAE~\cite{zheng2025rae}) suffer from under-corrupted signals at standard noise levels, impairing diffusion training. To compensate for our exceptionally high channel capacity, we adopt a dimension-dependent time-shifting schedule~\cite{pmlr-v235-esser24a}. The shift factor, mathematically defined relative to a base dimension of 4096, evaluates precisely to $\alpha = \sqrt{(C \times H \times W) / 4096} = \sqrt{(1024 \times 16^2) / 4096} = 8.0$. This properly calibrates the noise schedule, dynamically allocating representational capacity to high-curvature flow regions.

Finally, to enforce the manifold constraint, the neural network's unconstrained output is orthogonalized into tangent and radial components. A radial penalty $\lambda_{\text{rad}}=0.1$ is applied to strictly suppress non-tangential velocity predictions. Crucially, because our formulation relies heavily on sensitive geometric operations—such as evaluating trigonometric functions (sine, cosine, and arccosine) for spherical interpolations and computing explicit tangent projections—we strictly execute all forward computations and optimization steps in single-precision floating-point format (FP32). This mitigates the accumulation of truncation errors inherent in lower-precision formats (e.g., BF16) and guarantees the numerical stability required to accurately navigate the hyperspherical manifold.

\begin{algorithm}[h!]
\caption{Training Riemannian Flow Matching on HAE}
\label{alg:training}
\begin{algorithmic}[1]
\small 
\Require Pre-trained Encoder $\mathcal{E}$, DiT $v_\theta$, Radius $R=32.0$, Time-shift $\alpha=8.0$, Penalty $\lambda_{\text{rad}}=0.1$
\State Initialize DiT parameters $\theta$
\While{not converged}
    \State Sample image $\mathbf{x} \sim p_{\text{data}}$
    \State Extract and project latent: $\mathbf{z}_0 = R \cdot \frac{\mathcal{E}(\mathbf{x})}{\|\mathcal{E}(\mathbf{x})\|_2}$ \Comment{Spherical Projection}
    \State Sample and project noise: $\mathbf{z}_1 \sim \mathcal{N}(\mathbf{0}, \mathbf{I}), \quad \mathbf{z}_1 \leftarrow R \cdot \frac{\mathbf{z}_1}{\|\mathbf{z}_1\|_2}$
    \State Sample $t_{\text{base}} \sim \text{LogitNormal}(0, 1)$, shift time: $t \leftarrow \frac{\alpha \cdot t_{\text{base}}}{1 + (\alpha - 1) t_{\text{base}}}$
    
    \State \textbf{\textit{// Geodesic computations strictly evaluated in FP32}}
    \State Angular distance: $\Omega = \arccos\left(\langle \frac{\mathbf{z}_0}{R}, \frac{\mathbf{z}_1}{R} \rangle\right)$ \Comment{Geodesic SLERP setup}
    \State Interpolate state: $\mathbf{z}_t = R \left( \frac{\sin((1-t)\Omega)}{\sin(\Omega)}\frac{\mathbf{z}_0}{R} + \frac{\sin(t\Omega)}{\sin(\Omega)}\frac{\mathbf{z}_1}{R} \right)$
    \State Target velocity: $\mathbf{u}_t = R \frac{\Omega}{\sin(\Omega)} \left( \cos(t\Omega)\frac{\mathbf{z}_1}{R} - \cos((1-t)\Omega)\frac{\mathbf{z}_0}{R} \right)$
    
    \State Predict velocity $\mathbf{v} = v_\theta(\mathbf{z}_t, t)$ and normal vector $\bar{\mathbf{z}}_t = \frac{\mathbf{z}_t}{\|\mathbf{z}_t\|_2}$
    \State Target decomp.: $\mathbf{u}_{\text{rad}} = \langle \mathbf{u}_t, \bar{\mathbf{z}}_t \rangle \bar{\mathbf{z}}_t, \quad \mathbf{u}_{\text{tan}} = \mathbf{u}_t - \mathbf{u}_{\text{rad}}$
    \State Predict decomp.: $\mathbf{v}_{\text{rad}} = \langle \mathbf{v}, \bar{\mathbf{z}}_t \rangle \bar{\mathbf{z}}_t, \quad \mathbf{v}_{\text{tan}} = \mathbf{v} - \mathbf{v}_{\text{rad}}$
    \State Loss: $\mathcal{L} = \text{Mean}(\|\mathbf{v}_{\text{tan}} - \mathbf{u}_{\text{tan}}\|_2^2) + \lambda_{\text{rad}} \cdot \text{Mean}(\|\mathbf{v}_{\text{rad}}\|_2^2)$ \Comment{RFM + Penalty}
    \State Update $\theta$ using $\nabla_\theta \mathcal{L}$
\EndWhile
\State \Return Optimized parameters $\theta$
\end{algorithmic}
\end{algorithm}
\clearpage

\subsection{Inference Algorithm}
Algorithm~\ref{alg:inference} outlines the inference process utilizing the Riemannian Euler sampler. The sampling trajectory strictly adheres to the geometry of the hypersphere via the exponential map (Rodrigues' rotation formula). To maintain consistency with the training phase, the same time-shifting schedule ($\alpha=8.0$) is applied. At each step, we employ an Auto-Guidance mechanism (Classifier-Free Guidance); the model concurrently evaluates the conditioned and unconditioned states using a null token to extrapolate the guided velocity. The resulting velocity is then explicitly projected onto the tangent space before stepping forward to ensure manifold constraint satisfaction. Finally, before passing the generated latent to the pre-trained HAE decoder, we explicitly rescale the latent norm from the Flow Matching radius ($R=32.0$) to a designated decoding scale of $14.0$. As detailed in our ablation study (Appendix~\ref{sec:appendix_scale}), this rescaling step optimally calibrates the feature variance fed into the decoder, preventing numerical instability while preserving the network's capacity to synthesize intricate high-frequency details.

\begin{algorithm}[h!]
\caption{Inference with Riemannian Euler Sampler and Auto-Guidance}
\label{alg:inference}
\begin{algorithmic}[1]
\small 
\Require Trained DiT $v_\theta$, Pre-trained Decoder $\mathcal{D}$, Radius $R=32.0$, Steps $N$, Time-shift $\alpha=8.0$, Guidance scale $w$, Cond. $y$, Null token $\emptyset$, VAE scale $s_{\text{vae}}=14.0$
\State Sample noise $\mathbf{z}_1 \sim \mathcal{N}(\mathbf{0}, \mathbf{I})$ and project: $\mathbf{z}_1 \leftarrow R \cdot \frac{\mathbf{z}_1}{\|\mathbf{z}_1\|_2}$
\State Base time $t_{\text{base}} = \{1.0, 1.0 - \frac{1}{N}, \dots, 0.0\}$, shift time: $t_i = \frac{\alpha \cdot t_{\text{base}, i}}{1 + (\alpha - 1) t_{\text{base}, i}}$
\For{$i = 1$ to $N$}
    \State $\Delta t = t_{i+1} - t_i$ \Comment{Negative time step for reverse generation}
    
    \State \textbf{\textit{// Auto-Guidance (CFG) Step}}
    \State Predict unconditioned velocity: $\mathbf{v}_{\text{uncond}} = v_\theta(\mathbf{z}_{t_i}, t_i, \emptyset)$
    \State Predict conditioned velocity: $\mathbf{v}_{\text{cond}} = v_\theta(\mathbf{z}_{t_i}, t_i, y)$
    \State Apply Auto-Guidance: $\mathbf{v} = \mathbf{v}_{\text{uncond}} + w \cdot (\mathbf{v}_{\text{cond}} - \mathbf{v}_{\text{uncond}})$
    
    \State Project to tangent space: $\mathbf{v} \leftarrow \mathbf{v} - \langle \mathbf{v}, \frac{\mathbf{z}_{t_i}}{R} \rangle \frac{\mathbf{z}_{t_i}}{R}$ \Comment{Remove radial component}
    \State Compute rotation angle $\theta = \frac{|\Delta t| \cdot \|\mathbf{v}\|_2}{R}$ and direction $\bar{\mathbf{v}} = \text{sign}(\Delta t) \frac{\mathbf{v}}{\|\mathbf{v}\|_2 + \epsilon}$
    \State Exponential map: $\mathbf{z}_{\text{next}} = \mathbf{z}_{t_i} \cos(\theta) + R \bar{\mathbf{v}} \sin(\theta)$ \Comment{Rodrigues' rotation}
    \State Re-normalize: $\mathbf{z}_{t_{i+1}} = R \cdot \frac{\mathbf{z}_{\text{next}}}{\|\mathbf{z}_{\text{next}}\|_2}$
\EndFor
\State Rescale to target VAE scale $\mathbf{z}_0 \leftarrow s_{\text{vae}} \cdot \frac{\mathbf{z}_0}{R}$ and Decode $\hat{\mathbf{x}} = \mathcal{D}(\mathbf{z}_0)$
\State \Return $\hat{\mathbf{x}}$
\end{algorithmic}
\end{algorithm}

\section{Autoencoder Architecture and Training Details}
\label{sec:appendix_autoencoder}

In this section, we provide the detailed architectural configurations and hyperparameter settings for training our HAE framework. As discussed in Section~\ref{sec:method}, the training process is divided into multiple stages to balance semantic alignment and reconstruction fidelity. Autoencoder training was performed on 8 NVIDIA B200 GPUs and required approximately 1 hour and 20 minutes per epoch. Here, we detail the core architectures and the foundational configuration used in Stage 1.

\subsection{Hierarchical Convolutional Patch Embedding}
To mitigate the irreversible loss of high-frequency spatial details caused by the aggressive single-layer downsampling of standard ViTs, we replace the standard patch embedding with a Hierarchical Convolutional Stem. This module progressively downsamples the input image by a factor of 16 through four convolutional stages, smoothly projecting the 3-channel RGB input into the 1024-dimensional latent space required by the DINOv3-Large backbone. 

The exact architectural specifications are detailed in Table~\ref{tab:patch_embed_arch}. For the intermediate layers, we utilize Spatial Root Mean Square Normalization (TRMS2d) and SiLU activation functions to ensure stable gradient flow and robust feature extraction at the earliest stages.

\begin{table}[h!]
    \centering
    \caption{Architectural configuration of the Hierarchical Convolutional Patch Embedding.}
    \label{tab:patch_embed_arch}
            \resizebox{0.8\linewidth}{!}{%
    \begin{tabular}{lccccc}
        \toprule
        \textbf{Layer} & \textbf{Kernel Size} & \textbf{Stride} & \textbf{Output Channels} & \textbf{Norm} & \textbf{Activation} \\
        \midrule
        Input Image & - & - & 3 & - & - \\
        ConvLayer 1 & $7 \times 7$ & 2 & 64 & TRMS2d & SiLU \\
        ConvLayer 2 & $3 \times 3$ & 2 & 128 & TRMS2d & SiLU \\
        ConvLayer 3 & $3 \times 3$ & 2 & 256 & TRMS2d & SiLU \\
        ConvLayer 4 & $3 \times 3$ & 2 & 1024 & None & None \\
        \bottomrule
    \end{tabular}
    }
\end{table}

\subsection{Decoder Architecture}
We adopt the highly efficient decoder architecture proposed in DC-AE~\cite{chen2024deep}, tailored to our specific hyperspherical latent space. The decoder progressively upsamples the latent representation $\mathbf{z} \in \mathbb{R}^{H_z \times W_z \times 1024}$ back to the pixel space $\hat{\mathbf{x}} \in \mathbb{R}^{H \times W \times 3}$. The detailed layer-by-layer configuration is summarized in Table~\ref{tab:decoder_arch}. Across all blocks, we utilize TRMS2d and SiLU activation. For the upsampling stages, we employ InterpolateConv blocks with channel matching, utilizing a duplicating shortcut connection to preserve structural integrity.

\begin{table}[h!]
    \centering
    \caption{Architectural configuration of the HAE Decoder.}
    \label{tab:decoder_arch}
                \resizebox{0.7\linewidth}{!}{%
    \begin{tabular}{lcccc}
        \toprule
        \textbf{Stage} & \textbf{Width (Channels)} & \textbf{Depth (Blocks)} & \textbf{Block Type} \\
        \midrule
        Input Latent   & 1024  & - & - \\
        Stage 1 & 1024  & 3 & EViT-GLU \\
        Stage 2 & 1024  & 3 & EViT-GLU \\
        Stage 3 & 512   & 3 & ResBlock \\
        Stage 4 & 512   & 5 & ResBlock \\
        Stage 5 & 256   & 5 & ResBlock \\
        \midrule
        Output Image   & 3     & - & Conv2D (ReLU) \\
        \bottomrule
    \end{tabular}
    }
\end{table}

\subsection{Stage 1: Semantic-Structural Alignment Hyperparameters}
In Stage 1, we optimize the hierarchical convolutional patch embedding alongside the decoder to establish a stable semantic latent space. The model is trained using the AdamW optimizer with a base learning rate of $1 \times 10^{-5}$ for a total of 20 epochs. To ensure stable convergence, we employ Exponential Moving Average (EMA) and mixed-precision (BF16) training. The comprehensive hyperparameter configurations are detailed in Table~\ref{tab:stage1_hyper}.

\begin{table}[h!]
    \centering
    \caption{Hyperparameters for Stage 1 Autoencoder Training.}
    \label{tab:stage1_hyper}
    \begin{tabular}{lc}
        \toprule
        \textbf{Hyperparameter} & \textbf{Value} \\
        \midrule
        \multicolumn{2}{c}{\textit{Optimization Settings}} \\
        \midrule
        Optimizer & AdamW \\
        Learning Rate & $1 \times 10^{-5}$ \\
        Weight Decay & 0.0 \\
        Optimizer Betas & $(0.9, 0.999)$ \\
        Global Batch Size & 512 \\
        Precision & BF16 \\
        EMA Decay ($\beta_{\text{EMA}}$) & 0.9999 \\
        Total Epochs & 20 \\
        \midrule
        \multicolumn{2}{c}{\textit{Loss Objective Weights}} \\
        \midrule
        Reconstruction Weight ($\lambda_{\text{L1}}$) & 1.0 \\
        Perceptual Weight ($\lambda_{\text{lpips}}$) & 1.0 \\
        Cosine Similarity Weight ($\lambda_{\text{cos}}$) & 0.5 \\
        \bottomrule
    \end{tabular}
\end{table}

\subsection{Stage 2: Adversarial Adaptation with Stochastic Latent Smoothing}
Building upon the stable semantic space established in Stage 1, Stage 2 introduces adversarial training and stochastic latent smoothing to bridge the domain gap for downstream generative modeling. In this stage, we freeze the pre-trained DINOv3 encoder and train only the decoder alongside a feature-level discriminator (DINO-Disc). Following StyleGAN-T~\cite{sauer2023stylegan} and RAE~\cite{zheng2025diffusion}, this DINO-Disc performs discrimination in the feature space of a pre-trained DINO model to ensure logically consistent textures.

To ensure stable adversarial dynamics, we apply a two-stage warmup strategy. For the first two epochs, the discriminator is trained independently to adapt to the generated distribution, while the generator is optimized solely with reconstruction losses. From the third epoch onwards, the adversarial loss ($\lambda_{\text{adv}} = 0.5$) is fully activated. Specifically, we employ a hinge loss formulation for the adversarial objective ($\mathcal{L}_{\text{GAN}}$), updating the overall training loss as follows:
\begin{equation}
    \mathcal{L}_{\text{Stage2}} = \mathcal{L}_{\text{Stage1}} + \lambda_{\text{adv}}\mathcal{L}_{\text{GAN}}(\mathbf{x}, \hat{\mathbf{x}})
\end{equation}
Notably, we disable the cosine similarity loss ($\lambda_{\text{cos}} = 0.0$) in this stage, as the encoder is frozen and further directional constraints are unnecessary.

Crucially, we implement a progressive noise schedule for the stochastic latent smoothing. The spherical noise scale $\tau$ is initialized at $0.8$ and linearly increased to $1.2$ over the first 20 epochs, after which it remains constant until the end of training at epoch 40. This curriculum smoothing effectively thickens the latent manifold without abruptly collapsing the reconstruction capability. Both the generator (decoder) and the discriminator are optimized using AdamW with a reduced learning rate of $5 \times 10^{-5}$ and modified momentum betas of $(0.5, 0.9)$ to stabilize the GAN training. The complete hyperparameter settings are summarized in Table~\ref{tab:stage2_hyper}.

\begin{table}[h!]
    \centering
    \caption{Hyperparameters for Stage 2 Autoencoder Training (Adversarial Adaptation).}
    \label{tab:stage2_hyper}
    \begin{tabular}{lc}
        \toprule
        \textbf{Hyperparameter} & \textbf{Value} \\
        \midrule
        \multicolumn{2}{c}{\textit{Optimization Settings}} \\
        \midrule
        Optimizer (Generator \& Discriminator) \& AdamW \\
        Learning Rate & $5 \times 10^{-5}$ \\
        Weight Decay & 0.0 \\
        Optimizer Betas & $(0.5, 0.9)$ \\
        Global Batch Size & 512 \\
        Precision & BF16 \\
        EMA Decay ($\beta_{\text{EMA}}$) & 0.9975 \\
        Total Epochs & 40 \\
        Discriminator Warmup Epochs & 2 \\
        \midrule
        \multicolumn{2}{c}{\textit{Stochastic Latent Smoothing}} \\
        \midrule
        Noise Scale ($\tau$) Schedule & $0.8 \rightarrow 1.2$ (Epoch 0--20) \\
                                      & $1.2$ (Epoch 20--40) \\
        \midrule
        \multicolumn{2}{c}{\textit{Loss Objective Weights}} \\
        \midrule
        Reconstruction Weight ($\lambda_{\text{L1}}$) & 1.0 \\
        Perceptual Weight ($\lambda_{\text{lpips}}$) & 1.0 \\
        Adversarial Weight ($\lambda_{\text{adv}}$) & 0.5 \\
        Cosine Similarity Weight ($\lambda_{\text{cos}}$) & 0.0 \\
        \bottomrule
    \end{tabular}
\end{table}

\section{Ablation on Decoder Latent Scaling Factor}
\label{sec:appendix_scale}

As demonstrated in Section~\ref{sec:rfm}, our fully trained decoder exhibits strong robustness to latent magnitude shifts. However, during the image generation phase, the specific scaling factor $s_{\text{vae}}$ applied to the latents generated via Riemannian Flow Matching (RFM) significantly impacts the final visual quality. Because the RFM process outputs generative trajectories strictly on a hypersphere of radius $R=32.0$, these uncalibrated vectors must be properly rescaled before being passed to the decoder to synthesize the final image.

\begin{figure}[htbp]
    \centering
    \includegraphics[width=0.6\linewidth]{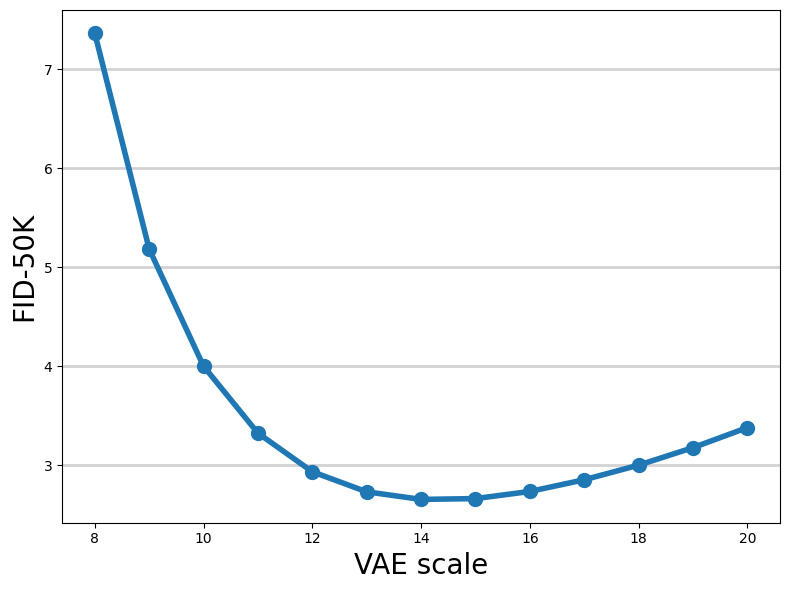} 
    \caption{Effect of the latent scaling factor ($s_{\text{vae}}$) on generative performance. The model achieves an optimal gFID balance at $s_{\text{vae}}=14.0$.}
    \label{fig:scale_ablation_metrics}
\end{figure}

To determine the optimal decoding scale for these generated latents, we conducted a systematic ablation study by sweeping the parameter $s_{\text{vae}}$ from 8.0 to 20.0 during the inference stage. As illustrated by the generative metrics in Figure~\ref{fig:scale_ablation_metrics}, the performance exhibits a clear U-shaped trajectory with respect to the scaling factor. Extremely low scaling factors excessively compress the latent variance, yielding a poor gFID of 7.36 at $s_{\text{vae}} = 8.0$. In this constrained regime, the restricted magnitude fails to sufficiently excite the decoder's feature maps, preventing the network from synthesizing intricate high-frequency details. 

\begin{figure}[htbp]
    \centering
    \includegraphics[width=1.0\linewidth]{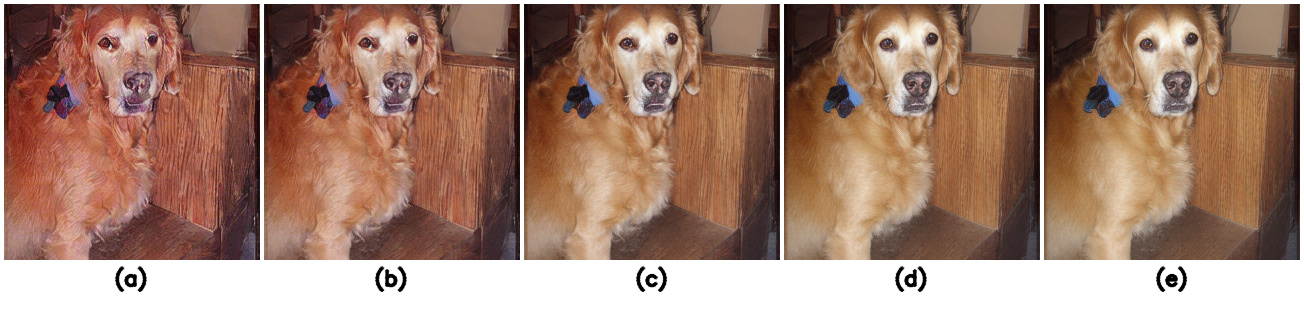} 
    \caption{Generated Golden Retriever samples across different scales: (a) 8.0, (b) 10.0, (c) 14.0, (d) 18.0, and (e) 20.0. As shown, $s_{\text{vae}}=14.0$ best preserves the intricate high-frequency details.}
    \label{fig:scale_ablation_vis}
\end{figure}

As the scale increases, the generative quality improves sharply, reaching a clear "sweet spot" at $s_{\text{vae}} = 14.0$ with an optimal gFID of 2.65. This qualitative progression is explicitly visualized in Figure~\ref{fig:scale_ablation_vis}. At lower scaling factors (e.g., $s_{\text{vae}}=8.0$ and $10.0$), the constrained feature magnitude prevents the decoder from forming coherent geometry, resulting in images that suffer from structural fragmentation and severe tearing artifacts. As the scale increases, the generative quality improves sharply, reaching a clear "sweet spot" at $s_{\text{vae}} = 14.0$. At this optimal scale, the model achieves the highest image quality, perfectly resolving fine textures and intricate high-frequency details. Conversely, excessively high scaling values (e.g., $s_{\text{vae}}=18.0$ and $20.0$) push the latent activations far beyond the decoder's learned optimal range. This disruption in feature statistics causes the decoding process to over-smooth the outputs, leading to a noticeable loss of texture and increasingly blurry images. Thus, we adopted $s_{\text{vae}} = 14.0$ as the default rescaling factor during inference to optimally align the generated latent statistics with the decoder's representational capacity.

\section{Limitations}
\label{sec:appendix_limitation}
While HAE demonstrates strong training efficiency and competitive generation quality, several limitations remain. First, our generative model has currently been trained for only 550 epochs, and we have not yet fully explored longer training regimes. Nevertheless, even at this intermediate training budget, HAE with LightningDiT-XL already achieves performance comparable to RAE with LightningDiT-XL trained for 800 epochs, suggesting favorable convergence efficiency. Second, our Riemannian Flow Matching training currently requires full-precision computation for stable optimization. In preliminary experiments, BF16 training led to numerical instability. As a result, our current implementation sacrifices some of the memory and throughput advantages typically provided by mixed-precision training. Finally, although our method substantially improves the efficiency of standard LightningDiT-XL on hyperspherical latents, it does not yet outperform the strongest specialized baselines such as RAE with DiT$^{\mathrm{DH}}$. In particular, DiT$^{\mathrm{DH}}$ uses a heavier architecture, and future work should investigate whether the proposed hyperspherical formulation can be combined with such stronger transformer heads or further architectural scaling.

\section{Uncurated Generated Samples}
\label{sec:appendix_samples}

In this section, we present uncurated $256 \times 256$ samples generated by LightningDiT-XL to provide a comprehensive view of its generative performance. All images were generated using the Riemannian Euler sampler for 50 steps with a Classifier-Free Guidance (CFG) scale of 1.1. 

\begin{figure*}[h!]
    \centering
    \captionsetup{font=small}
    
    \begin{minipage}[t]{0.48\textwidth}
        \centering
        \includegraphics[width=\linewidth]{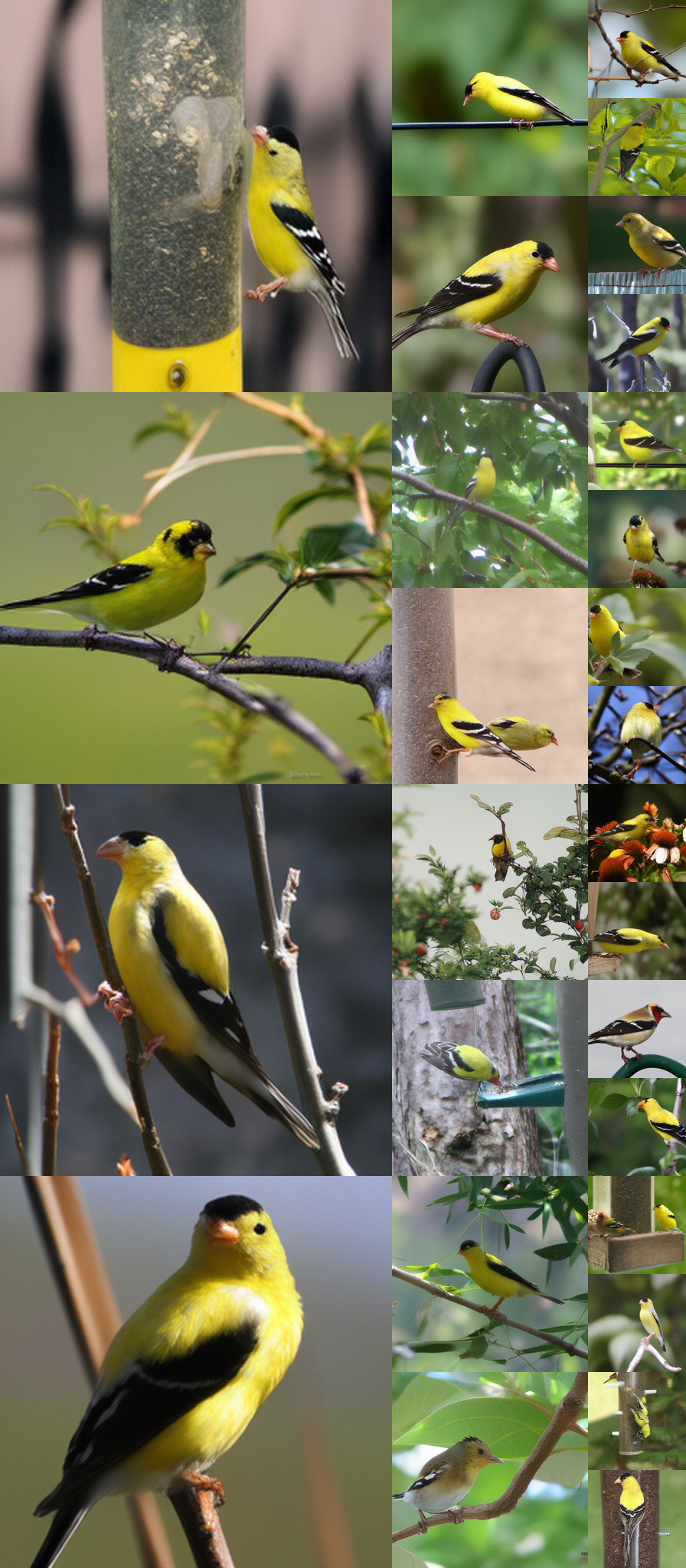}
        \caption{Uncurated $256 \times 256$ samples of class Goldfinch at CFG 1.1.}
        \label{fig:uncurated_dog}
    \end{minipage}
    \hfill
    \begin{minipage}[t]{0.48\textwidth}
        \centering
        \includegraphics[width=\linewidth]{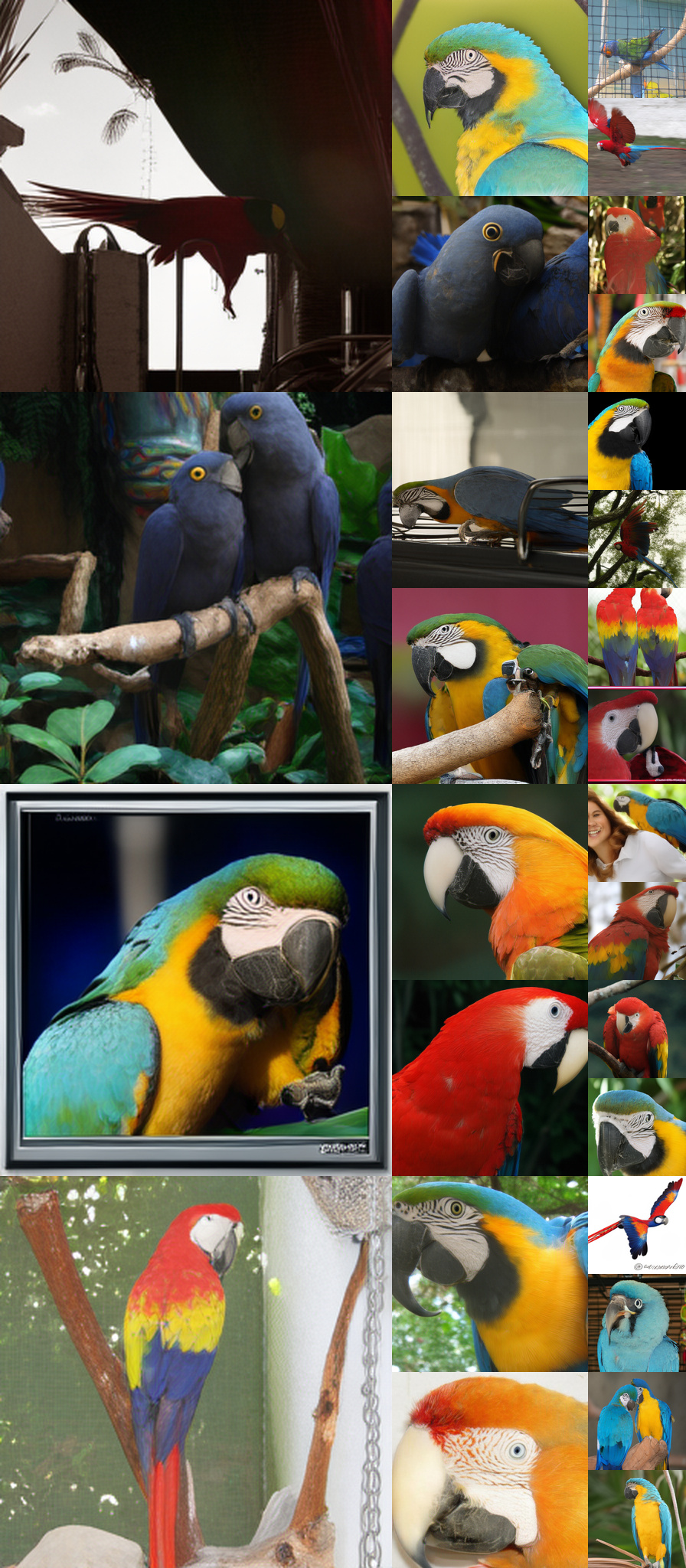}
        \caption{Uncurated $256 \times 256$ samples of class Macaw at CFG 1.1.}
        \label{fig:uncurated_macaw}
    \end{minipage}
\end{figure*}

\begin{figure*}[t]
    \centering
    \captionsetup{font=small}
    
    \begin{minipage}[t]{0.48\textwidth}
        \centering
        \includegraphics[width=\linewidth]{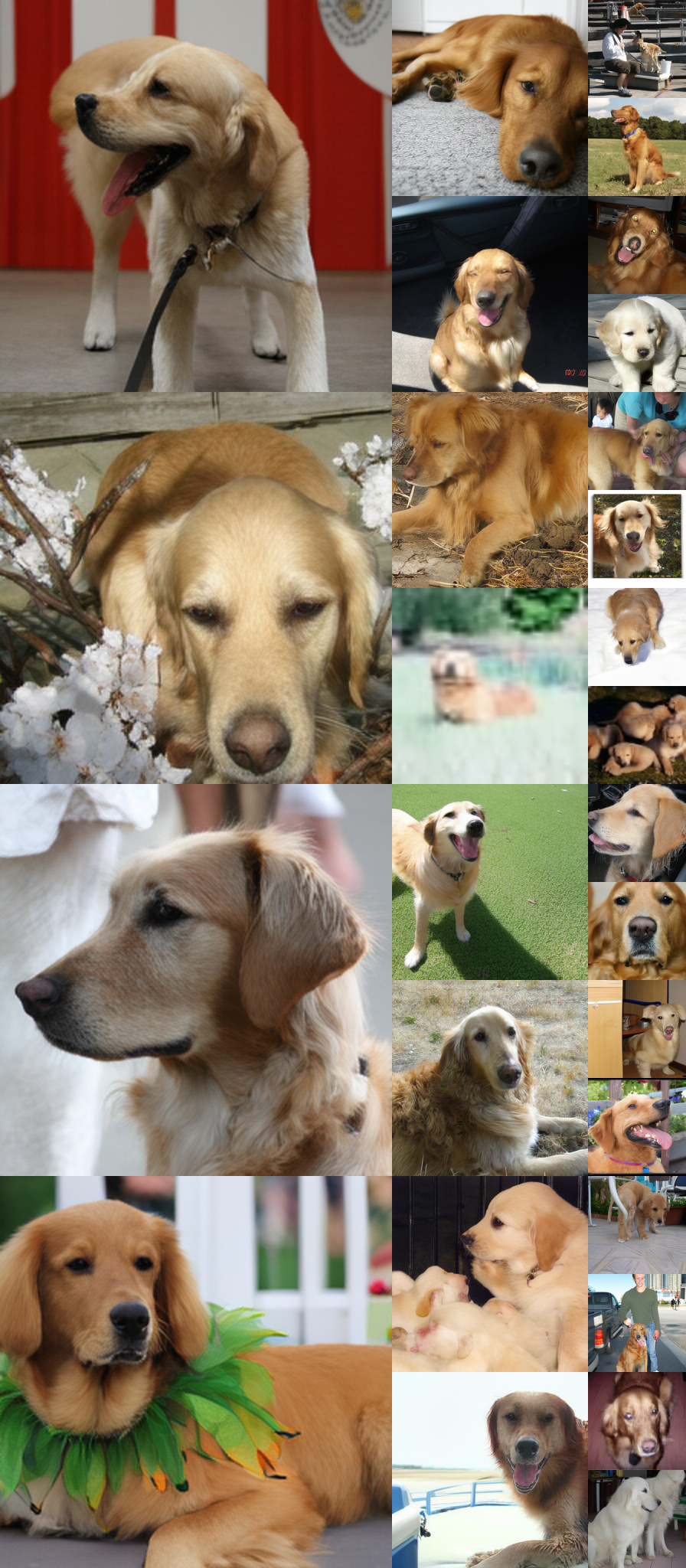}
        \caption{Uncurated $256 \times 256$ samples of class Golden Retriever at CFG 1.1.}
        \label{fig:uncurated_dog}
    \end{minipage}
    \hfill
    \begin{minipage}[t]{0.48\textwidth}
        \centering
        \includegraphics[width=\linewidth]{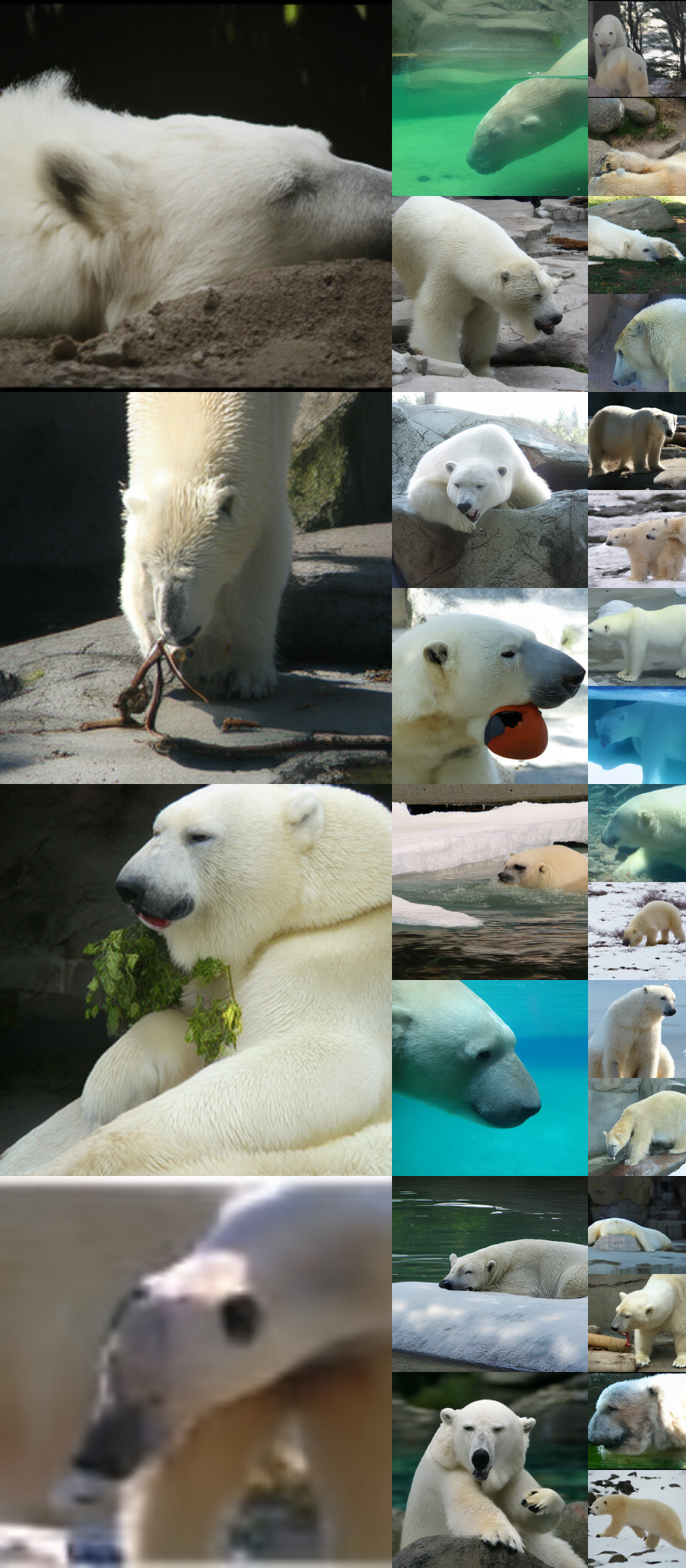}
        \caption{Uncurated $256 \times 256$ samples of class Polar bear  at CFG 1.1.}
        \label{fig:uncurated_macaw}
    \end{minipage}
\end{figure*}

\begin{figure*}[t]
    \centering
    \captionsetup{font=small}
    
    \begin{minipage}[t]{0.48\textwidth}
        \centering
        \includegraphics[width=\linewidth]{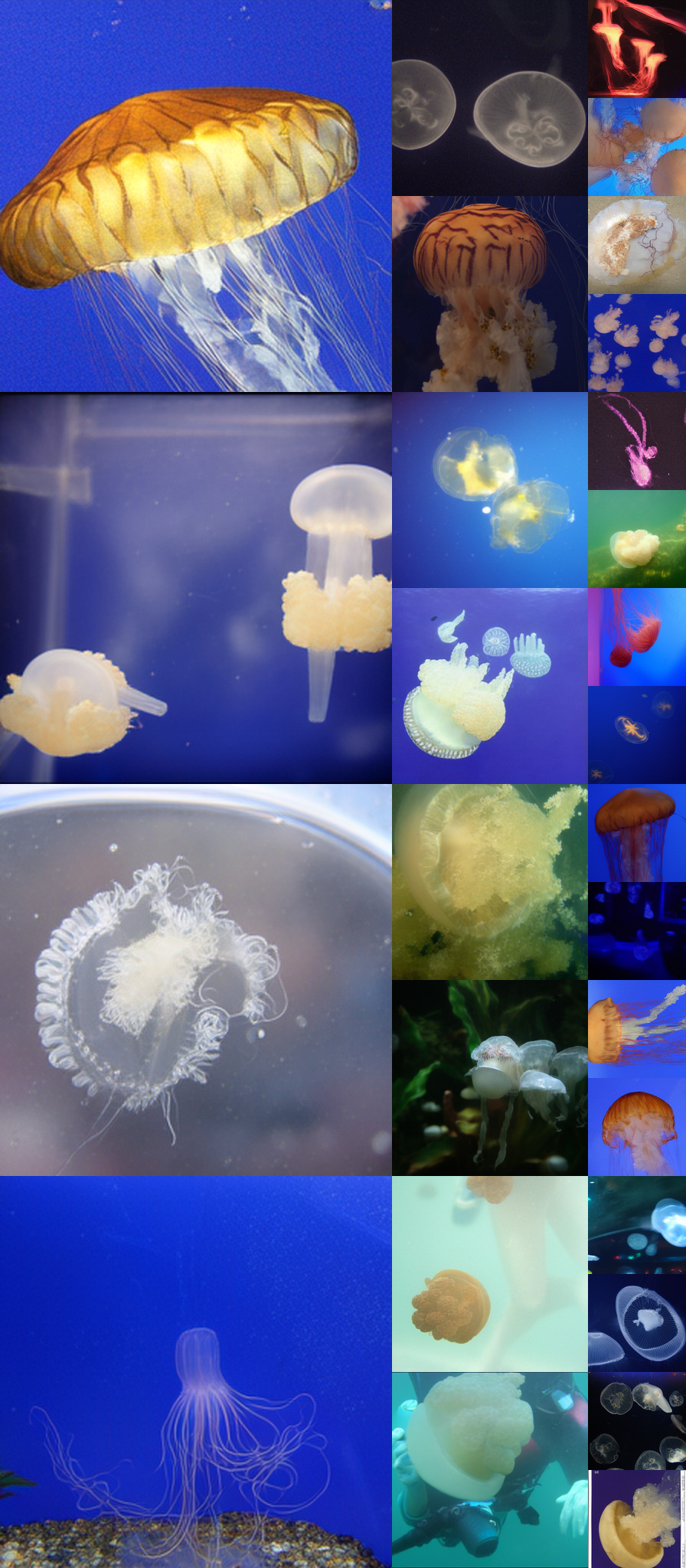}
        \caption{Uncurated $256 \times 256$ samples of class Jellyfish  at CFG 1.1.}
        \label{fig:uncurated_dog}
    \end{minipage}
    \hfill
    \begin{minipage}[t]{0.48\textwidth}
        \centering
        \includegraphics[width=\linewidth]{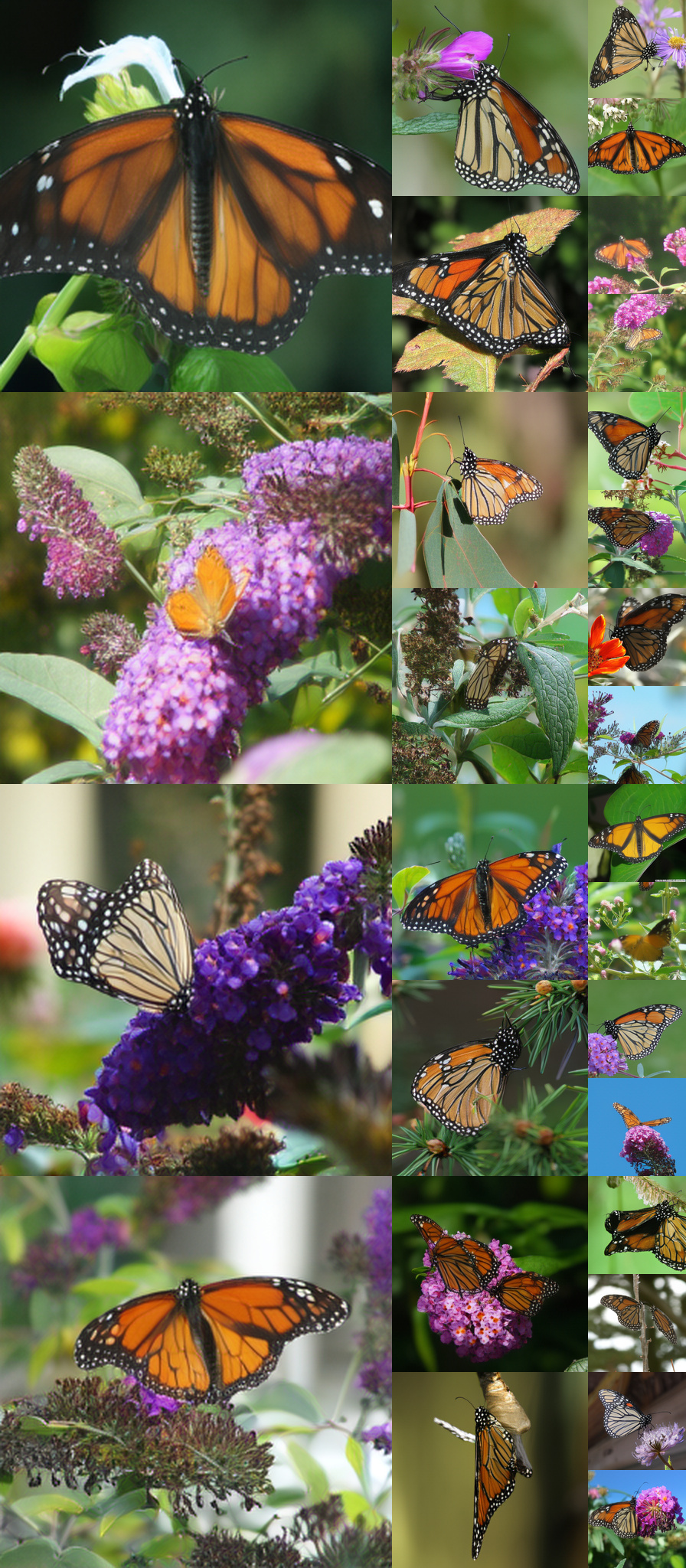}
        \caption{Uncurated $256 \times 256$ samples of class Monarch butterfly  at CFG 1.1.}
        \label{fig:uncurated_macaw}
    \end{minipage}
\end{figure*}

\begin{figure*}[t]
    \centering
    \captionsetup{font=small}
    
    \begin{minipage}[t]{0.48\textwidth}
        \centering
        \includegraphics[width=\linewidth]{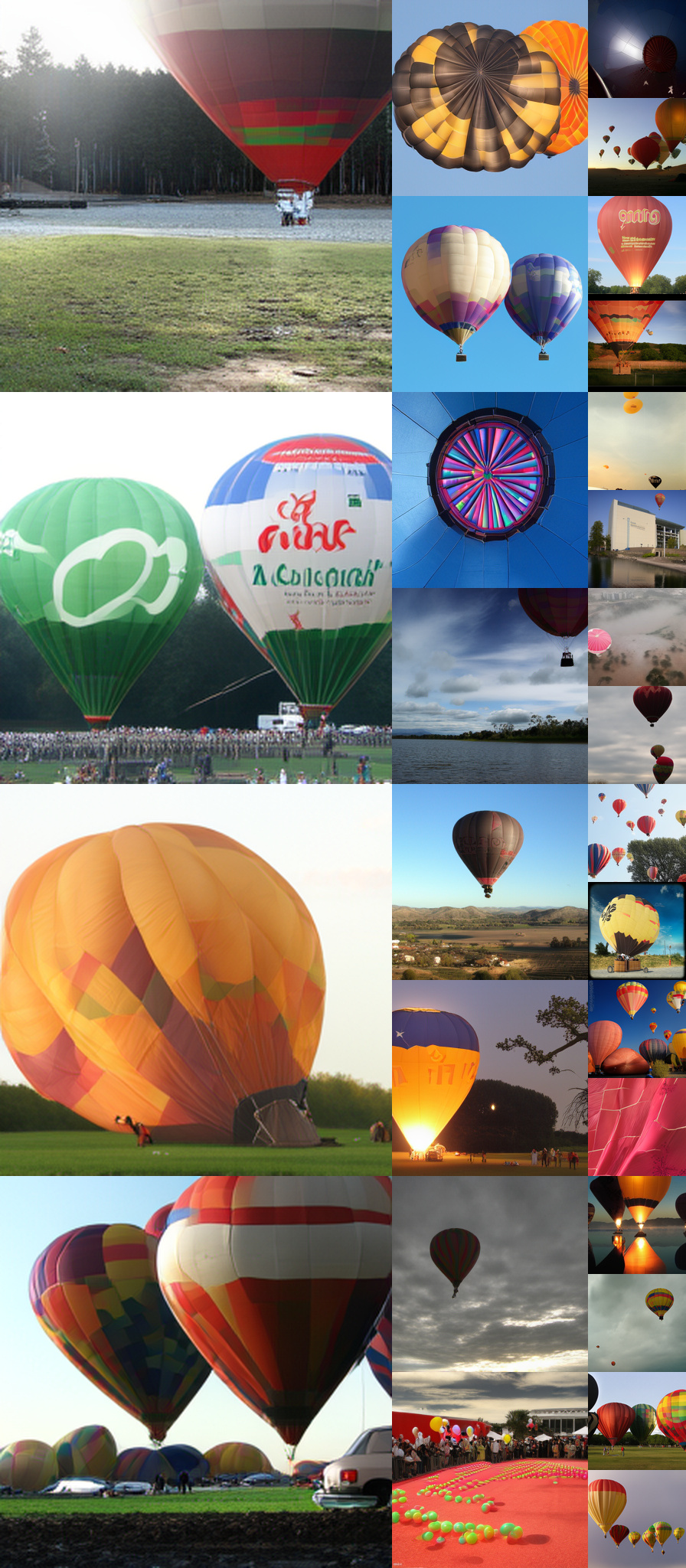}
        \caption{Uncurated $256 \times 256$ samples of class Balloon  at CFG 1.1.}
        \label{fig:uncurated_dog}
    \end{minipage}
    \hfill
    \begin{minipage}[t]{0.48\textwidth}
        \centering
        \includegraphics[width=\linewidth]{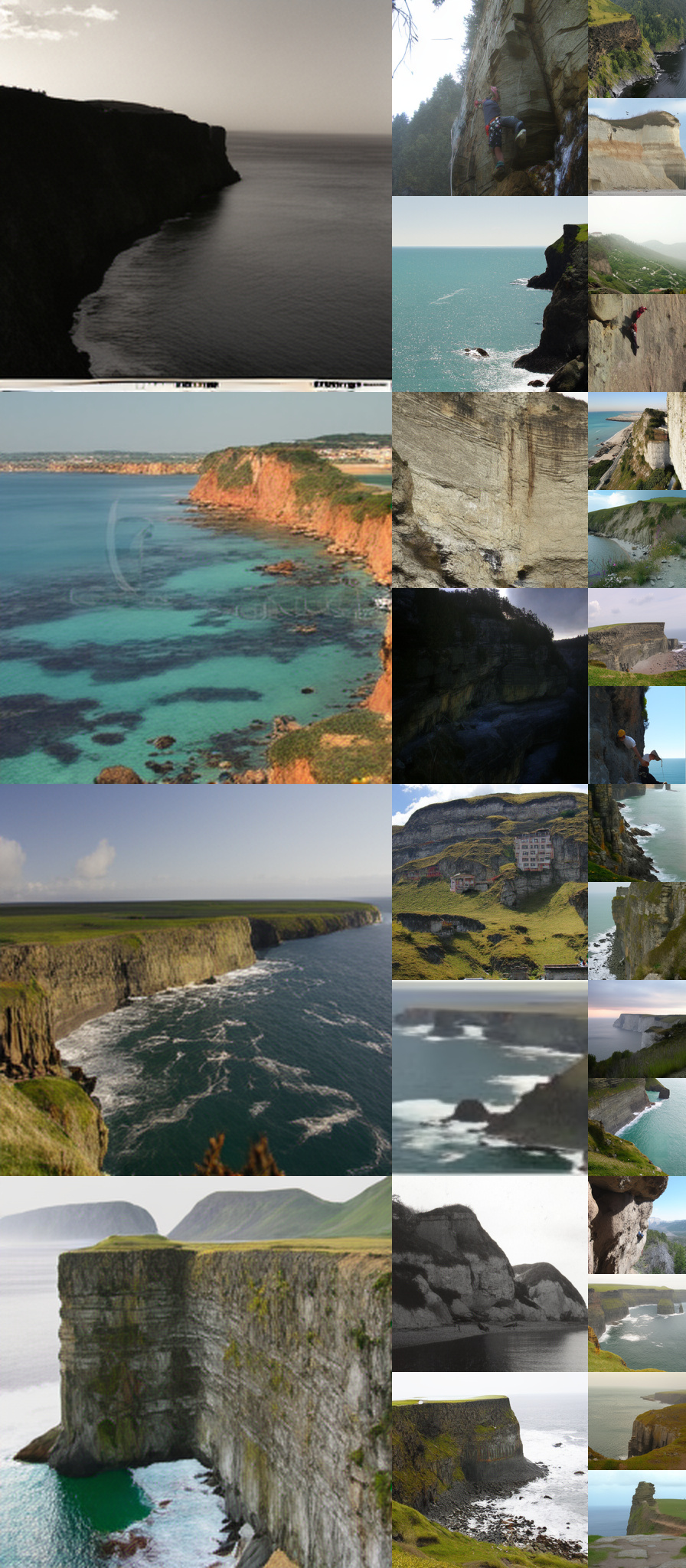}
        \caption{Uncurated $256 \times 256$ samples of class Cliff  at CFG 1.1.}
        \label{fig:uncurated_macaw}
    \end{minipage}
\end{figure*}

\begin{figure*}[t]
    \centering
    \captionsetup{font=small}
    
    \begin{minipage}[t]{0.48\textwidth}
        \centering
        \includegraphics[width=\linewidth]{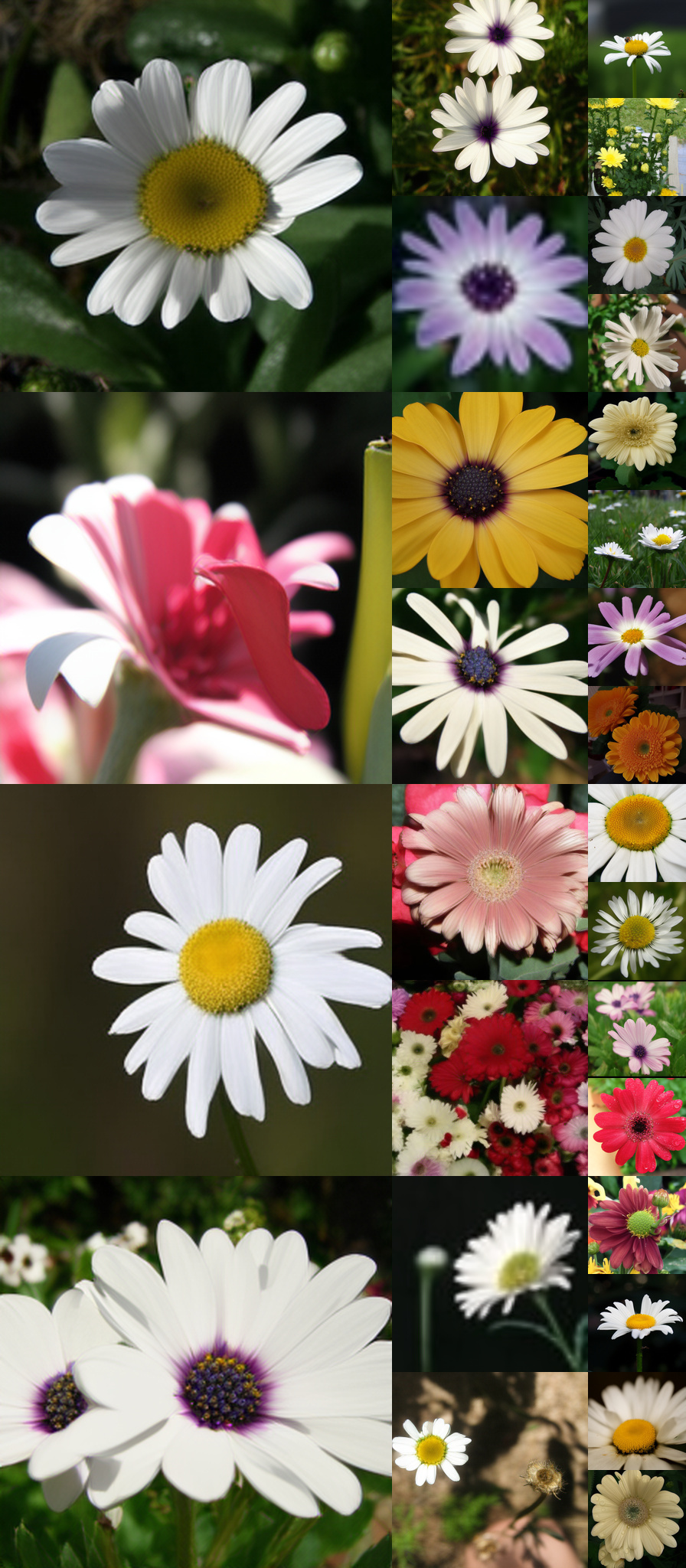}
        \caption{Uncurated $256 \times 256$ samples of class Daisy  at CFG 1.1.}
        \label{fig:uncurated_dog}
    \end{minipage}
    \hfill
    \begin{minipage}[t]{0.48\textwidth}
        \centering
        \includegraphics[width=\linewidth]{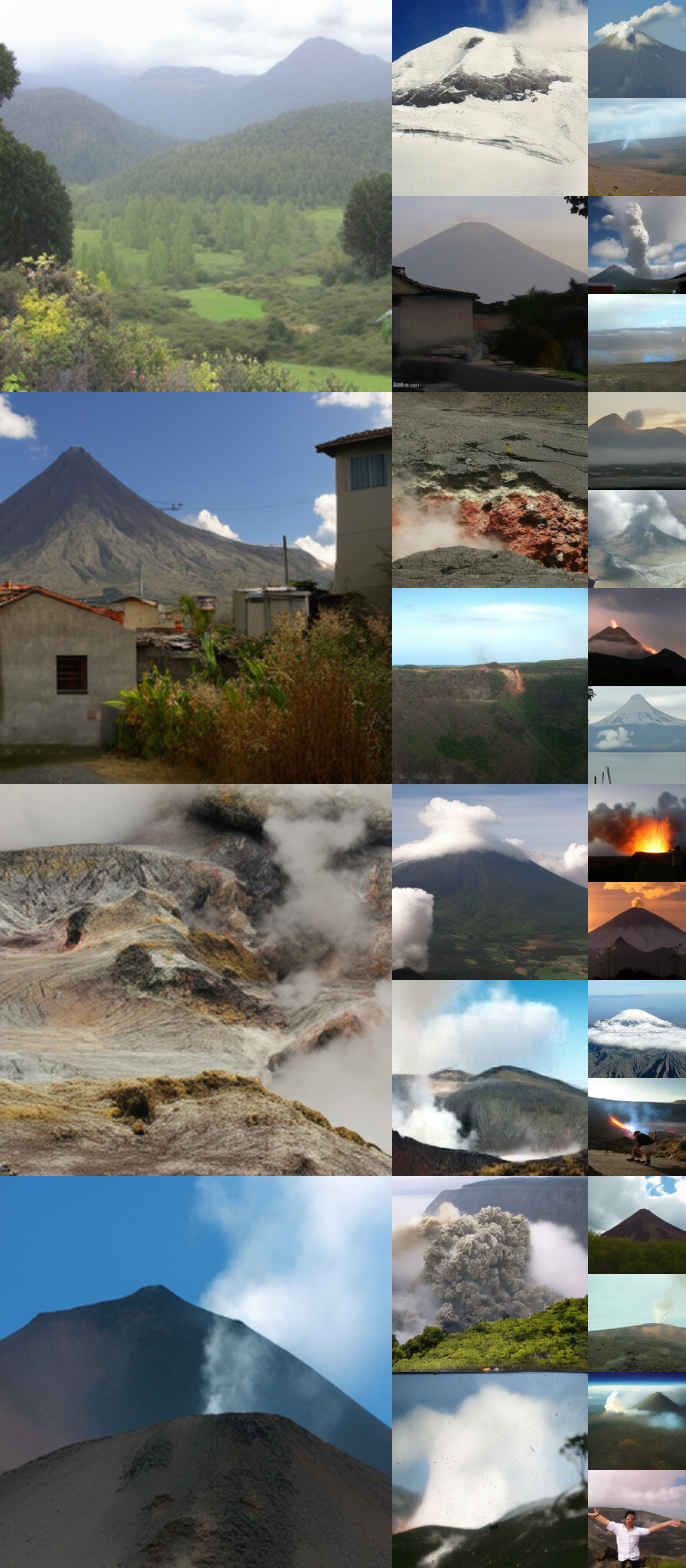}
        \caption{Uncurated $256 \times 256$ samples of class Volcano  at CFG 1.1.}
        \label{fig:uncurated_macaw}
    \end{minipage}
\end{figure*}

\begin{figure*}[t]
    \centering
    \captionsetup{font=small}
    
    \begin{minipage}[t]{0.48\textwidth}
        \centering
        \includegraphics[width=\linewidth]{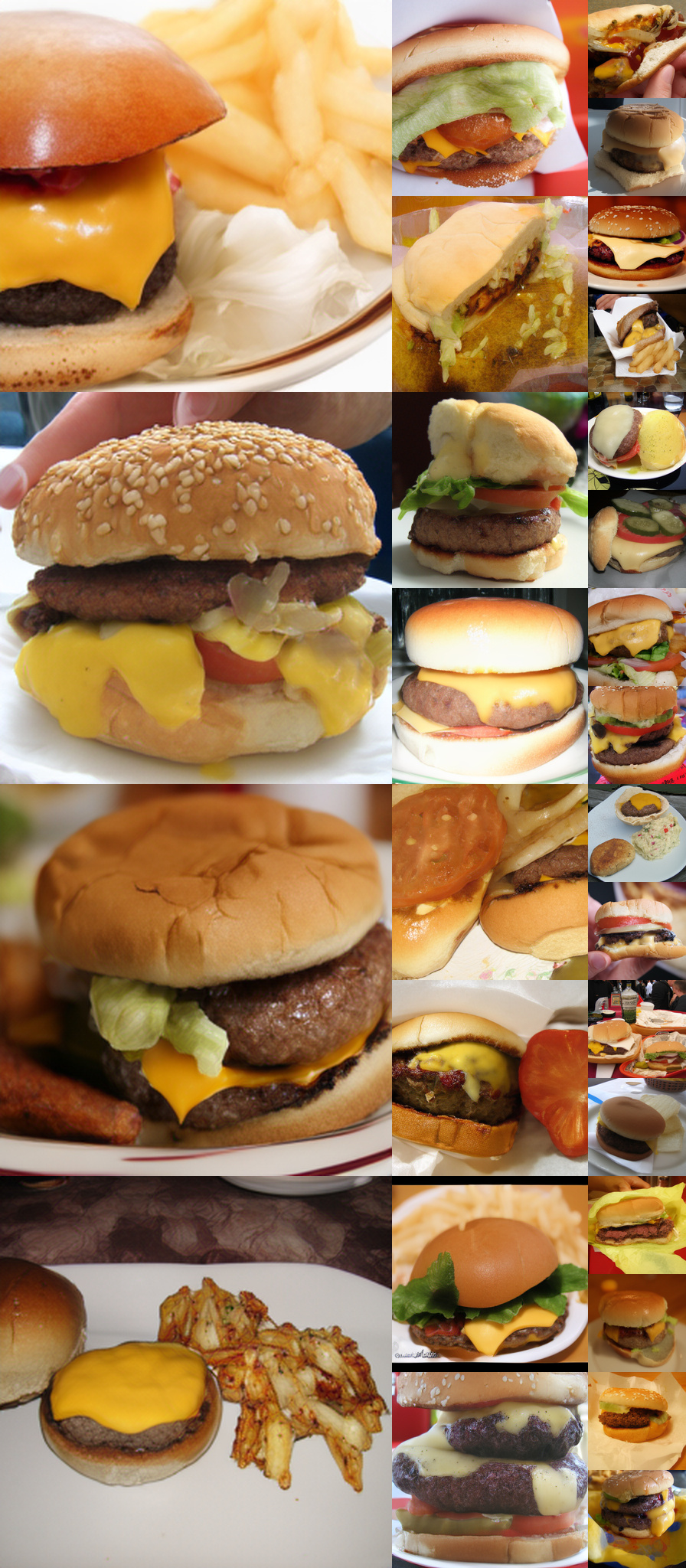}
        \caption{Uncurated $256 \times 256$ samples of class Hamburger  at CFG 1.1.}
        \label{fig:uncurated_dog}
    \end{minipage}
    \hfill
    \begin{minipage}[t]{0.48\textwidth}
        \centering
        \includegraphics[width=\linewidth]{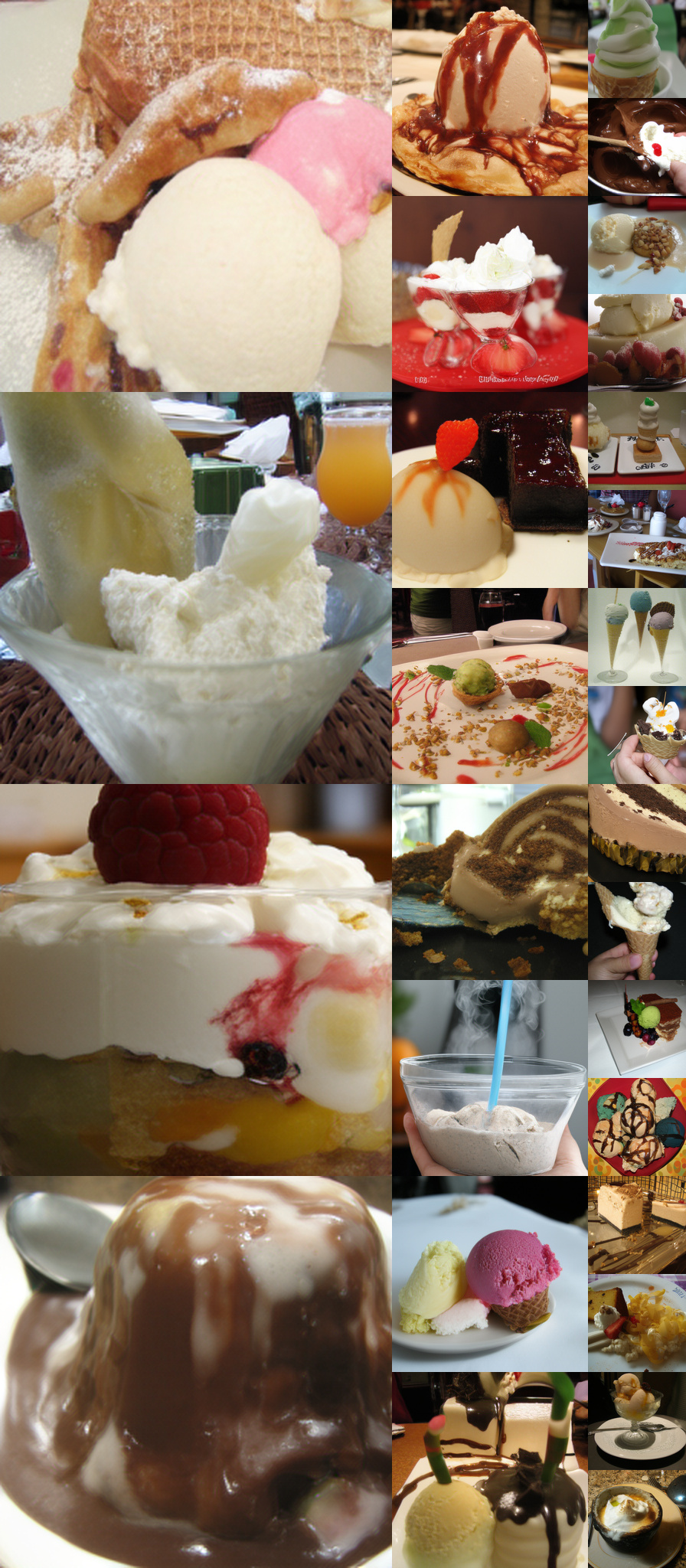}
        \caption{Uncurated $256 \times 256$ samples of class Icecream  at CFG 1.1.}
        \label{fig:uncurated_macaw}
    \end{minipage}
\end{figure*}

\clearpage

\end{document}